\documentclass{article}

\PassOptionsToPackage{numbers, compress}{natbib}
\usepackage[dblblindworkshop, final]{neurips_2025}

\workshoptitle{NeurIPS 2025 Workshop on Generative AI in Finance}% the "default" option is equal to the "main" option, which is used for the Main Track with double-blind reviewing.
\usepackage{graphicx}
\usepackage[utf8]{inputenc} % allow utf-8 input
\usepackage[T1]{fontenc}    % use 8-bit T1 fonts
\usepackage{hyperref}       % hyperlinks
\usepackage{url}            % simple URL typesetting
\usepackage{booktabs}       % professional-quality tables
\usepackage{amsfonts}       % blackboard math symbols
\usepackage{nicefrac}       % compact symbols for 1/2, etc.
\usepackage{microtype}      % microtypography
\usepackage{xcolor}         % colors
\usepackage{amsmath}
\usepackage{amsthm}
\newtheorem{theorem}{Theorem}[section]
\newtheorem{corollary}{Corollary}[theorem]
\newtheorem{proposition}[theorem]{Proposition}
\usepackage{subcaption}
\usepackage{amssymb}
\usepackage{placeins}
\usepackage{multirow}
\usepackage{enumerate}   

\title{Multi-Trajectory Physics-Informed Neural Networks for HJB Equations with Hard-Zero Terminal Inventory: Optimal Execution on Synthetic \& SPY Data}

\author{%
  Anthime Valin\thanks{This work was conducted in partial fulfillment of the requirements for the MSc in Financial Technology at University College London.} \\
  Department of Computer Science \\
  University College London\\
  \texttt{anthime.valin@gmail.com} \\
}

\begin{document}

\maketitle

\begin{abstract}
We study optimal trade execution with a hard-zero terminal inventory constraint, modeled via Hamilton-Jacobi-Bellman (HJB) equations. Vanilla PINNs often under-enforce this constraint and produce unstable controls. We propose a Multi-Trajectory PINN (MT-PINN) that adds a rollout-based trajectory loss and propagates a terminal penalty on \(X_T\) via backpropagation-through-time, directly enforcing \(X_T=0\). A lightweight \(\lambda\)-curriculum is adopted to stabilize training as the state expands from a risk-neutral reduced HJB to a risk-averse HJB. On the Gatheral-Schied single-asset model \citep{gatheral2011optimal}, MT-PINN aligns closely with their derived closed-form solutions and concentrates terminal inventory tightly around zero while reducing errors along optimal paths. We apply MT-PINNs on SPY intraday data, matching TWAP when risk-neutral, and achieving lower exposure and competitive costs, especially in falling windows, for higher risk-aversion. 

\end{abstract}

\section{Introduction}

Many financial decision-making problems, ranging from optimal portfolio and consumption choice to optimal order execution, can be written as optimal control problems, with the value function governed by a Hamilton-Jacobi-Bellman (HJB) PDE \citep{pham2009continuous}. In optimal trade execution, a necessary constraint is hard-zero terminal inventory (i.e, all shares must be liquidated by the end of the horizon). This requirement creates a singularity near terminal (\(\tau =T-t \rightarrow 0)\), where the value function exhibits non-smooth structure. Vanilla PINNs, trained on PDE residuals and soft boundary penalties, frequently under-enforce this constraint \citep{wang2022respecting}, often yielding non-zero inventory in simulated rollouts and unstable control near maturity.

We propose a Multi-Trajectory PINN (MT-PINN) that directly penalizes terminal inventory via a rollout-based trajectory loss, using backpropagation-through-time (BPTT) to propagate the terminal penalty through simulated trajectories (see Appendix \ref{Appendix B}). This decreases terminal inventory violations and reduces tv error along optimal control paths. A lightweight \(\lambda\)-curriculum (from risk-neutral to risk-averse) is employed to further stabilize training. We evaluate MT-PINNs' performance on a single-asset execution model, originally proposed by Gatheral and Schied in \citep{gatheral2011optimal}, demonstrating its capabilities in aligning with the closed-form solution trading rates and concentrating terminal inventory tightly around zero compared to vanilla PINNs. Furthermore, we apply MT-PINNs to SPY intraday data, showing risk aversion explicitly reflected in the expected exposure-cost trade-off, coinciding with TWAP when risk neutral (\(\lambda=0\)) and revealing advantages over TWAP in falling markets. Specifically, we make the following contributions: (i) a trajectory-aware loss that enforces \(X_T = 0\); (ii) a simple curriculum regularization for risk aversion; and (iii) synthetic + real-market validation with strong constraint satisfaction and stable controls.

\section{Problem Setup and Method}

\paragraph{Model \& notation.}
We study a variant of the Almgren-Chriss optimal execution model, proposed by Gatheral and Schied \citep{gatheral2011optimal}, where the unaffected price \(S_t\) follows GBM (\(dS_t = \sigma S_t dW_t\), for volatility \(\sigma > 0\)) with market impact comprising linear permanent impact with strength \(\kappa > 0\) and standard temporary impact. We work over a horizon \(T > 0\) with time-to-maturity \(\tau = T -t\) and initial inventory \(X_0 > 0\). Inventory \(X_t\) evolves under the sell trading rate \(v_t\) via \(\dot{X}_t = -v_t\) with a hard terminal constraint \(X_T = 0\). The value function is denoted by \(\Gamma(\tau, X, S)\) in the risk-averse case and \(\Gamma(\tau, X)\) in the risk-neutral case. It represents the minimum expected cost from the current state to the terminal, accounting for market impact and, when applicable, risk aversion \(\lambda\). 

\paragraph{Reduced HJB.}
The HJB equation takes different forms in the risk-neutral and risk-averse regimes (the risk-neutral case is $S$-invariant). Minimizing the Hamiltonian with respect to $v$ and substituting the optimal feedback $v^*=\frac{1}{2}\partial \Gamma/\partial X$, yields the reduced HJB:
\[
\frac{\partial \Gamma}{\partial \tau} =
\begin{cases}
\kappa^2 X^2 - \dfrac{1}{4}\Big(\dfrac{\partial \Gamma}{\partial X}\Big)^2, & \lambda=0, \ \ \text{(Proposition \ref{prop:OE-HJB-zero})} \\
\dfrac{1}{2}\sigma^2 S^2 \dfrac{\partial^2 \Gamma}{\partial S^2} + \kappa^2 X^2 + \lambda S X - \dfrac{1}{4}\Big(\dfrac{\partial \Gamma}{\partial X}\Big)^2, & \lambda>0 \ \ \text{(Proposition \ref{prop:OE-HJB-pos})}.
\end{cases}
\]
with terminal condition \(\Gamma(\tau \to 0,\, X, S) = 0 \) if \(X=0\) and \(+\infty\) otherwise. At \(\lambda =0\) the value is price-invariant, so the state dimension is 1 (variables \((\tau, X)\)), while for \(\lambda > 0\) it is 2 (variables \((\tau, X, S)\)).

\paragraph{MT-PINN parameterization.}
We approximate the value function with a smooth MLP \(\hat{\Gamma}(\tau, X,S; \theta)\) (independent of \(S\) when \(\lambda =0\)). Derivatives found in the HJB equation are calculated using autodiff and the feedback control used for rollouts is \(v^*=\tfrac{1}{2}\,\partial \hat{\Gamma}/\partial X\).

\paragraph{Multi-trajectory terminal-inventory loss.}
To enforce the hard constraint \(X_T=0\), we add a rollout-based loss that penalizes terminal inventory across a batch of starting states and horizons. We fix the sets of initial inventories and prices $\{(X^{(p)}_0, S^{(p)}_0 ) \}^P_{p=1} \subset \mathcal{X}_0 \times \mathcal{S}_0$ and horizons $\{ T_j\}^J_{j=1} \subset (0, T]$. For each pair $(p,j)$, define $x^{(p)}_{T_j}$ as the terminal inventory obtained by rolling out the inventory dynamics under the network-implied control, using a forward Euler discretization with \(N_\text{dt}\) steps: 
\[X_{k+1} = X_k - v^*(\tau_k, X_k, S_k)\Delta t, \ \ \tau_k = T_j - t_k, \ \ \Delta t =\frac{T_j}{N_\text{dt}}.\]
The loss averages a composite penalty, \(\psi\), across the batch:

\[
\mathcal{L}_\text{traj} =  \frac{1}{PJ} \sum^P_{p=1} \sum^J_{j=1} \psi(x_{T_j}^{(p)}), \ \ \ \psi(x_T) = \begin{cases} |x_T|,& |x_T| \le 1, \\ x^2_T,& |x_T|  > 1. \end{cases}
\]
For \(\lambda = 0\) the rollouts depend only on \( \left ( X^{(p)}_0, T_j \right )\); for \(\lambda > 0\) they depend on \(\left ( X^{(p)}_0, S^{(p)}_0,T_j \right )\). Gradients are computed via backpropagation-through-time (BPTT) (see Appendix \ref{Appendix B}).

\paragraph{Composite loss.}
We then minimize a composite loss:
\[\mathcal{L}_\text{total}(\theta;\lambda) = w_\text{PDE}\mathcal{L}_\text{PDE}(\theta;\lambda)+w_\text{traj}\mathcal{L}_\text{traj}(\theta;\lambda)+w_\text{IC}\mathcal{L}_\text{IC}(\theta;\lambda)+w_\text{sym}\mathcal{L}_\text{sym}(\theta;\lambda) + \mathbf{1}_{\{\lambda>0\}} w_0 \mathcal{L}_\text{0-term}(\theta),\]

where \(\mathcal{L}_\text{PDE}\) is the squared residual of the reduced HJB above, \(\mathcal{L}_\text{IC}\) enforces the inventory-axis condition, \(\mathcal{L}_\text{sym}\) enforces the model's symmetry (see Corollary \ref{col:OE-sym-0} and Corollary \ref{col:OE-sym-lam}), and \(\mathcal{L}_\text{0-term}\) sets \(\Gamma(0,0,S) = 0\) for \(\lambda > 0\). Loss weights use DWA-style adaptive weighting (see Appendix \ref{Appendix C.2}).

\paragraph{\(\lambda\)-curriculum.} 
Following \citep{krishnapriyan2021characterizing}, we apply a curriculum regularization to the risk-aversion parameter $\lambda$. We choose a monotone schedule \(0=\lambda_0 < \lambda_1 < \cdot \cdot\cdot < \lambda_k = \lambda^*\) and, at stage \(k\), minimize \(\mathcal{L}_\text{total}(\theta;\lambda_k)\), warm-starting from the previous stage (\(\theta^{(k)} \leftarrow \theta^{k-1}\)). Training begins at \(\lambda_0 = 0\), where the value function is price-invariant, and then transitions to \(\lambda > 0\). This curriculum has been shown to stabilize training and yield lower PDE residuals \citep{krishnapriyan2021characterizing, huang2022hompinns}. 

\section{Synthetic Benchmark}
\paragraph{Baselines.}
We compare MT-PINN against two vanilla PINNs that omit the multi-trajectory term and instead use a quadratic terminal penalty \footnote{Empirically, it was found that if neither multi-trajectory nor this quadratic terminal penalty is used, the PINNs will converge to sub-optimal local minima.}: (i) Vanilla PINN, trained with HJB residual, terminal value penalty, and same structural terms as MT-PINN); (ii) Vanilla PINN + \(\lambda\)-curriculum, with the same loss as (i), but trained with the staged \(\lambda\) schedule as defined in \S2.

\paragraph{Setup.}
Fixed horizon \(T=5.0\); inventory domain \(X \in [-10, 10]\) and \(S \in [10, 100]\); Volatility \(\sigma =0.1\); permanent-impact strength \(\kappa =0.1\); Collocation points per run \(N_\text{PDE}=30,000\); For \(\lambda >0\), baselines and MT-PINN trained on 55k epochs; target risk aversion \(\lambda^* \in \{0, 0.05, 0.1\}\) with five curriculum stages with \(\lambda_\alpha = \alpha \cdot \lambda^*\), for \(\alpha \in (0.25, 0.50, 0.75, 0.9, 1.0)\) (used for  MT-PINN and baseline (ii)). The multi-trajectory batch samples \(P=820\) initial state \(\{(X_0^{(0)}, S^{(p)}_0)\}^P_{p=1}\) (omitting \(S^{(p)}_0\) when \(\lambda=0\)) and rolls out to horizons \(\{T/50, T/10, T/5, 2T/5, 3T/5, 4T/5, T\}\) with 200 Euler steps. All remaining details are provided in Appendix \ref{Appendix D}.

\begin{figure}
    \centering
    \includegraphics[width=\linewidth]{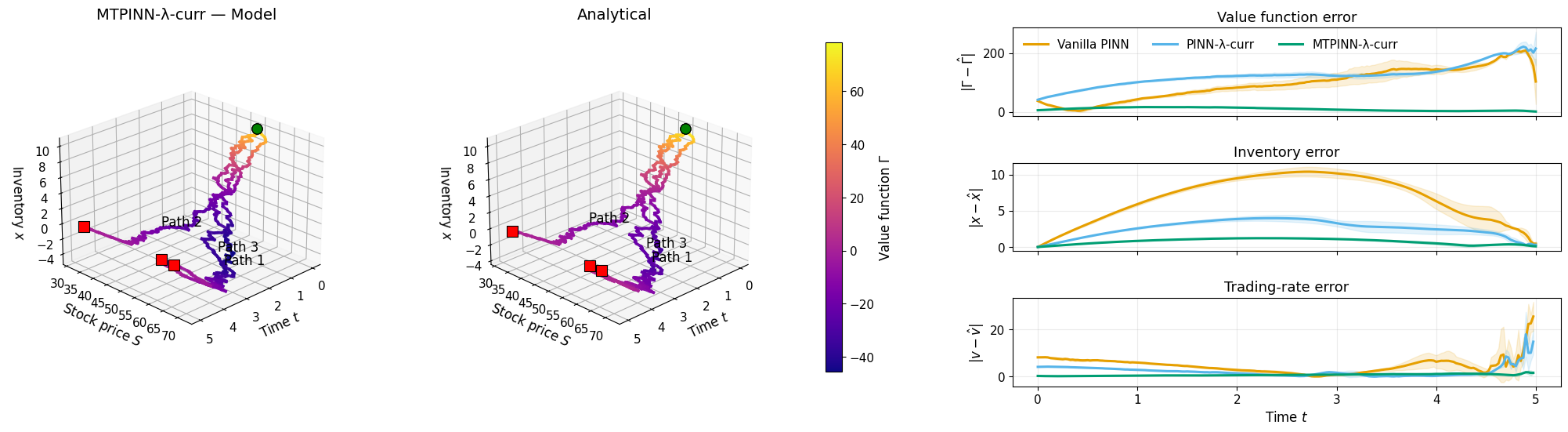}
    \caption{\textit{Left:} MT-PINN vs analytical optimal paths for \(\lambda=0.10\) on three price paths, colored by the value function \(\Gamma\) with a shared colormap. \textit{Right:} three stacked panels show mean \(\pm\) one standard deviation across the same paths of absolute errors along the optimal path: (top) \(|\Gamma - \hat{\Gamma}|\), (middle) \(|x -\hat{x}|\), (bottom) \(|v-\hat{v}|\).}
    \label{fig:paths}
\end{figure}

\paragraph{Key Results.}
%MT-PINN tracks the analytical control very closely across three price paths (Figure \ref{fig:trading-rate}), with errors remaining small throughout the horizon and without the late-maturity instability seen in vanilla PINNs. 
MT-PINN attains the lowest error along the optimal path across absolute value function error, inventory error, and trading rate error, clearly outperforming both baselines over most of the horizon and closely tracking the analytical control without the late-maturity instability seen in the baseline PINNs (see Figure \ref{fig:paths}). This translates into tighter enforcement of the hard-zero terminal constraint as evidenced in Table \ref{tab:terminal-enforcement}. Additional metrics, residual maps, and results comparing varying \(\lambda \in [0, 0.05, 0.10]\) are provided in Appendix \ref{Appendix F}.

%\begin{figure}
%    \centering\includegraphics[width=\linewidth]{Synthetic/trading_rate.png}
%  \caption{Optimal trading rate $v^*(t)$: model vs. closed form for $\lambda=0.10$ for three price paths}
%  \label{fig:trading-rate}
%\end{figure}

\begin{table}
  \caption{Absolute terminal-inventory enforcement on the synthetic benchmark for 200 simulated price paths and \(\lambda = 0.10\) (for varying \(\lambda\) see Table \ref{tab:terminal-enforcement-all-lam} in Appendix \ref{Appendix F}). We report $p_\varepsilon=\Pr(|X_T|\le\varepsilon)$ with $\varepsilon=0.05$.}
  \renewcommand{\arraystretch}{0.9}
  \label{tab:terminal-enforcement}
  \centering
  \begin{tabular}{lccc}
    \toprule
    & \multicolumn{2}{c}{$|X_T|$ statistics} & \\
    \cmidrule(r){2-3}
    Method & Mean $\pm$ Std & 95th pct & $p_\varepsilon$ \\
    \midrule
    Vanilla PINN           & 0.777 $\pm$ 0.444 & 1.407 & 0.055 \\
    PINN-\(\lambda\)-curr  & 0.164 $\pm$ 0.161 & 0.527 & 0.205 \\
    MT-PINN-\(\lambda\)-curr & \textbf{0.073 $\pm$ 0.092} & \textbf{0.241} & \textbf{0.600} \\
    \bottomrule
  \end{tabular}
\end{table}
\section{Real-Market SPY Backtest}

\begin{figure}
    \centering\includegraphics[width=\linewidth]{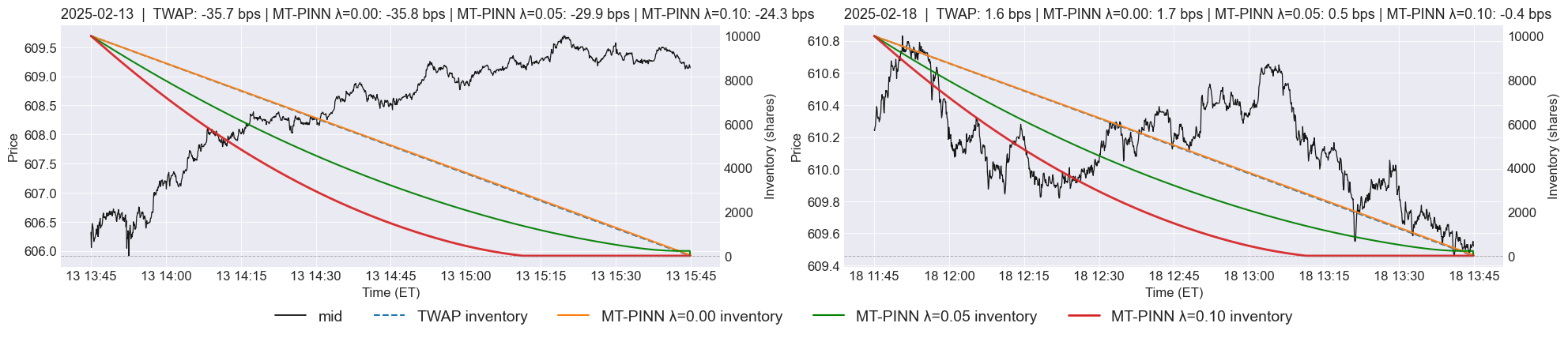}
  \caption{Execution trajectories and costs for MT-PINN with varying \(\lambda \in [0, 0.05, 0.1]\) versus TWAP, over two sampled trading windows of the SPY. \textit{Left:} rising SPY window; \textit{Right:} falling SPY window.}
  \label{fig:windows}
\end{figure}

\begin{table}[t]
  \caption{Execution exposure and cost comparison of MT\,-PINN (varying $\lambda$) vs.\ TWAP.
  Results averaged over $n=21$ windows. Costs in basis points (1 bps $=0.01\%$).}
  \label{tab:cost_variance}
  \centering
  \begin{tabular}{lcccc}
    \toprule
    \textbf{Model} & \textbf{Mean Exposure} & \textbf{Exposure Std.} & \textbf{Mean Cost (bps)} & \textbf{Cost Std. (bps)} \\
    \midrule
    TWAP                   & 0.334 & 0.0000 & -6.35 & 12.56 \\
    MT-PINN $\lambda=0.00$ & 0.336 & 0.0000 & -6.37 & 12.58 \\
    MT-PINN $\lambda=0.05$ & 0.231 & 0.0004 & -5.01 & 11.02 \\
    MT-PINN $\lambda=0.10$ & 0.164 & 0.0002 & -3.69 &  9.67 \\
    \bottomrule
  \end{tabular}
\end{table}

\paragraph{Setup.}
The trading day on SPY mid-price is split into three 2-hour fixed intraday windows on 5-second intervals, and repeated across seven days (Feb 10 - Feb 19, 2025, excluding weekends)\footnote{Windows exclude the first 15 minutes of market open and the last 15 minutes of market close. This was done to avoid the noisy and elevated volatility typically observed at those times \citep{INCI201779}.}, and assumes no trading fees or overnight risk. Uses the same Gatheral-Schied dynamics/penalties as \S2. MT-PINN with \(\lambda \in \{0, 0.05, 0.10\}\) is compared against TWAP; \(\lambda=0\) serves as the risk-neutral MT-PINN which should coincide with TWAP in expectation. Time horizon is expressed by trading day units, where elapsed time is normalized by the effective length of a trading day on the U.S. equity markets (i.e., \(T=0.308\)). Inventory is normalized by the initial position, so that the inventory lies in the interval \([-1, 1]\), and represents the fraction of the initial inventory. Stock price \(S \in [590, 620]\) is set based on the mid-price range over the full 21 time windows. Daily realized volatility is calculated from the data and is set to \(\sigma = 0.0038\) (\(\approx 6\%\) annualized), and permanent-impact strength is set to \(\kappa=0.2\). The MT-PINN setup follows similarly to \S3. For more details see Appendix \ref{Appendix E}.

\paragraph{Key Results.}
MT-PINN with \(\lambda=0\) closely matches TWAP, both in exposure  (0.336 vs. 0.334) and cost (-6.37 vs. -6.35 bps), confirming the risk-neutral case (see Table \ref{tab:cost_variance}). Increasing risk aversion traces a clean risk-cost frontier: mean exposure drops while average cost rises moderately (likely due to the general uptrend of the SPY). Figure \ref{fig:windows} shows \(\lambda>0\) front-loads the execution and outperforms TWAP in down-moves, whereas MT-PINN (with \(\lambda=0\))/TWAP remains competitive in rising windows. Across all windows, MT-PINN policies remain smooth and strongly satisfy the zero terminal inventory constraint, aligning with the synthetic benchmark (see Appendix \ref{Appendix F} for more results).

\paragraph{Reproducibility.}
Code, configs, and scripts to reproduce all figures and experiments are provided at: \url{https://github.com/anthimevalin/Multi-Trajectory-PINNs-Zero-Terminal-HJB}.

\section{Conclusion \& Limitations}
MT-PINN enforces hard-zero terminal inventory via a rollout-based trajectory loss and, with a simple $\lambda$-curriculum, produces stable controls and less error along the optimal path. On the Gatheral–Schied benchmark it matches the closed-form control and concentrates $X_T$ tightly around zero; on SPY intraday data, $\lambda=0$ behaves like TWAP, while $\lambda>0$ traces a clear risk–cost frontier and performs best in falling windows. 
\paragraph{Limitations.} The impact model used doesn't include fees, nonlinear temporary impact, and assumes frictionless execution beyond impact terms; experiments are single-asset and don't investigate cross-asset; and the trajectory penalty has per-epoch time complexity \(\mathcal{O}(J \times N_\text{dt} \times P)\) and memory complexity \(\mathcal{O}(N_\text{dt} \times P)\) (more details in Appendix \ref{Appendix B}).
Moreover, the execution model abstracts away market microstructure: execution occurs in a continuous-price setting without modeling order book depth, bid–ask dynamics, or order-flow imbalance. Consequently, the learned policies may overlook short-term liquidity constraints and queue dynamics that affect realized execution costs. Incorporating order-book-derived state variables would be a natural extension toward more realistic liquidity modeling.

\bibliographystyle{unsrtnat}
\bibliography{references}
\newpage
\appendix

\section{Mathematical details \& proofs} \label{Appendix A}

We seek the optimal strategy to liquidate an initial inventory of $X > 0$ shares over some fixed time horizon $[0, T]$. The unaffected asset price follows a driftless geometric Brownian motion (GBM):

\[ dS_t = \sigma S_t dW_t, \ \ \ S_0 > 0 \]

Gatheral and Schied in \citep{gatheral2011optimal} assume that the number of shares is represented by an absolutely continuous adapted process $X_t$ that evolves as:

\[ X_t = X - \int^t_0 v_s ds, \]

with boundary conditions $X_0 = X$ and $X_T = 0$ and where $v_t = \dot{X}_t \in \mathbb{R}$ is the trading rate at time $t$. Furthermore, transactions occur at 
\[ \tilde{S}_t = S_t + \eta v_t + \gamma(X_t - X_0),\]

where $S_t$ is the unaffected stock price process, $\eta v_t$ models temporary impact, and $\gamma (X_t - X_0)$ models permanent impact. Following \citep{gatheral2011optimal}, the temporary impact coefficient is scaled to one and the effect of permanent impact is represented by a quadratic inventory penalty with parameter \(\kappa\).

The control problem minimizes the expected cumulative cost

\[J =  \mathbb{E}\left[ \int^T_0 (v^2_t + \kappa^2 X^2_t + \lambda X_t S_t ) dt \right], \]

where \(\kappa > 0\) weights inventory-holding, \(\lambda \ge 0\) denotes the risk aversion parameter (VaR-based penalty on inventory exposure), and \(v^2_t\) encodes the temporary impact costs.

We formulate two risk-aversion regimes in the HJB equation: when $\lambda = 0$ and when $\lambda > 0$. 

\begin{proposition}[Risk-Neutral HJB] \label{prop:OE-HJB-zero}
For zero risk penalty (i.e., $\lambda = 0$), the resulting stochastic control problem, whose value function $\Gamma(\tau, X)$ with $\tau := T-t$ satisfies the nonlinear HJB equation under optimal control
\begin{align}
    \frac{\partial \Gamma}{\partial \tau} = \kappa^2 X^2 - \frac{1}{4}\left (\frac{\partial \Gamma}{\partial X} \right)^2 \label{eq:OE-HJB-0}
\end{align}
with

\begin{align}
    \Gamma(\tau,0) = 0,  \ \ \forall \tau \in [0,T] \label{eq:OE-BC-0}
\end{align} 
subject to the terminal condition

\begin{align}\label{OE-term-0}
    \Gamma(0, X) =
    \begin{cases}
        0, & \text{if } X=0 \\
        +\infty, & \text{otherwise.}
    \end{cases}
\end{align}
\end{proposition}
\begin{proof}
    The cost functional for zero risk penalty is

    \begin{align*}
         J(\tau, X) = \mathbb{E}\left[ \int^\tau_0 (v^2_t + \kappa^2 X^2_t)dt\right],
    \end{align*}
    and thereby no longer dependent on the price $S_t$. The value function can then be written as the infimum of the cost functional
    \begin{align*}
        \Gamma (\tau, X) = \inf_{v \in \mathbb{R}} \mathbb{E} \left [ \int ^\tau_0((v^2_t + \kappa^2 X^2_t)dt)\right],
    \end{align*}
Using the Principle of Optimality, we can represent the value function as:
    \begin{equation}
    \Gamma(\tau, X) = \inf_{v_{[t, t+\Delta t]}} \int^{t+\Delta t}_t  (v^2_s + \kappa^2 X_s^2)ds + \Gamma(\tau-\Delta t, X(t+\Delta t)) \label{eq:OE-HJB-0-proof}
    \end{equation}
    For small $\Delta t$: 
    \[\int^{t+\Delta t}_t (v^2_s + \kappa^2 X_s^2)ds = (v^2_t + \kappa^2 X_t^2)\Delta t + \mathcal{O}(\Delta t^2), \ \ \text{and}\]
    \begin{align*}
     X(t+\Delta t) &= X + \dot{X} \Delta t + \mathcal{O}(\Delta t^2)  \\
    &= X - v_t \Delta t + \mathcal{O}(\Delta t^2)
    \end{align*}
    
    Using Taylor expansion, the value function can be expressed by:
    
    \[\Gamma(\tau - \Delta t, X - v_t\Delta t) = \Gamma(\tau, X) - \frac{\partial \Gamma}{\partial \tau}\Delta t - v \frac{\partial \Gamma}{\partial X}\Delta t + \mathcal{O}(\Delta t^2)\]
    
    Inserting these back into \ref{eq:OE-HJB-0-proof} and subtracting $\Gamma(\tau,X)$ from both sides:
    
    \[0 = \inf_{v \in \mathbb{R}} \left [v^2 + \kappa^2 X^2 - \frac{\partial \Gamma}{\partial \tau} - v \frac{\partial \Gamma}{\partial X} \right]\Delta t+\mathcal{O}(\Delta t^2)\]
    
    Dividing by $\Delta t$ and letting $\Delta t \rightarrow 0$:
    
    \[ \frac{\partial \Gamma}{\partial \tau} =  \kappa^2 X^2 + \inf_{v \in \mathbb{R}} \left [v^2 - v \frac{\partial \Gamma}{\partial X} \right ]\]

The optimal control is calculated by minimizing the Hamiltonian $H(\frac{\partial \Gamma}{\partial X},v ) := v^2 - v\frac{\partial \Gamma}{\partial X}$ by setting its derivative with respect to $v$ to zero. This yields the optimal trading rate with respect to the value function:
\begin{equation}
    v^* = \frac{1}{2} \frac{\partial \Gamma}{\partial X} \label{eq:execution_rate}
\end{equation}

This further reduces the HJB to obtain (\ref{eq:OE-HJB-0})
\[
    \frac{\partial \Gamma}{\partial \tau} = \kappa^2 X^2 - \frac{1}{4}\left (\frac{\partial \Gamma}{\partial X} \right)^2 
\]

Note that the cost functional is minimized when implementing the "do nothing" control (i.e., $X_t = 0$), which results in the condition \ref{eq:OE-BC-0}:

\[\Gamma(\tau, 0) = 0, \ \ \ \forall \tau \in [0,T]\]
\end{proof}

\begin{proposition}[Risk-Averse HJB]
For positive risk penalty (i.e., $\lambda > 0$), the resulting stochastic control problem, whose value function $\Gamma(\tau, X,S)$ with $\tau := T-t$ satisfies the nonlinear HJB equation under optimal control \label{prop:OE-HJB-pos}

\begin{align}
    \frac{\partial\Gamma}{\partial \tau} = \frac{1}{2} \sigma^2 S^2     \frac{\partial^2 \Gamma}{\partial S^2} + \kappa^2 X^2 + \lambda S X - \frac{1}{4}  \left (\frac{\partial \Gamma}{\partial X} \right)^2, \label{eq:OE-HJB-lam}
\end{align}
with
\begin{align}
    \Gamma(\tau,0,S) \le 0 \quad \forall \tau \in [0,T], \ \forall S > 0, \label{eq:OE-BC-lam} 
\end{align} 
subject to terminal condition

\begin{align}\label{OE-term-lam}
    \Gamma(0, X,S) =
    \begin{cases}
        0, & \text{if } X=0 \\
        +\infty, & \text{otherwise}
    \end{cases}
\end{align}

\end{proposition}
\begin{proof}
    Gatheral and Schied in \citep{gatheral2011optimal} derive the HJB equation in equation (3.18) as
\begin{align}
    \frac{\partial\Gamma}{\partial \tau} = \frac{1}{2} \sigma^2 S^2     \frac{\partial^2 \Gamma}{\partial S^2} + \kappa^2 X^2 + \lambda S X + \inf_{v \in \mathbb{R}} \left(v^2-v \frac{\partial \Gamma}{\partial X} \right ), 
\end{align}

Reducing the HJB using \ref{eq:execution_rate}, we obtain \ref{eq:OE-HJB-lam}. To obtain \ref{eq:OE-BC-lam}, suppose a deterministic $y \in H^1_0 (0, \tau)$ with non-zero integral and such that $y(0) = y(\tau) = 0$, e.g., $y(t) = \sin(\frac{\pi t}{\tau})$. For $\epsilon > 0$ set:

\[X^{(\epsilon)}_t := \epsilon y(t), \ \ \ v^{(\epsilon)}_t := \dot{X}^{(\epsilon)}_t = \epsilon \dot{y}(t)\]
As $X^{(\epsilon)}_0 = X^{(\epsilon)}_\tau = 0$, $X^{(\epsilon)} \in \mathbb{R}$. Its cost functional is

\[J(\tau, X, S) = \mathbb{E} \left[ \int^\tau_0 \left( \left (v_t^{(\epsilon)} \right)^2 + \kappa^2 \left (X_t^{(\epsilon)}\right)^2 + \lambda X^{(\epsilon)}_t S_t\right)dt \right]\]
As $X^{(\epsilon)}$ is deterministic and $\mathbb{E} [S_t] = S_0$ for driftless GBM:

\[\mathbb{E} \left [ \int^\tau_0 \lambda X^{(\epsilon)}_t S_t dt\right] = \lambda S_0 \epsilon \int^\tau_0 y(t)dt,\]
such that

\[J(\tau, X, S) = \underbrace{\epsilon^2 \int^\tau_0 \dot{y}(t)^2 + \kappa^2 y(t)^2dt}_{:= A\epsilon^2} + \underbrace{\lambda S_0 \epsilon \int^\tau_0 y(t) dt}_{:= B \epsilon}\]

From Proposition \ref{prop:OE-HJB-zero}, we know $A \epsilon^2 \ge 0$. Choose sign of y so that $B \le 0$.  Then for sufficiently small $\epsilon \in [0, -B/2A]$:

    \[J(\tau, X, S) = A\epsilon^2  + B\epsilon \le 0, \ \ \therefore \ \Gamma(\tau, 0, S) \le 0\]

\end{proof}

Gatheral and Schied in \citep{gatheral2011optimal} showed that the closed-form optimal strategy is:

\[X^*_t = \sinh{((T-t)\kappa)} \left [ \frac{X}{\sinh(T \kappa)} - \frac{\lambda}{2\kappa} \int^t_0 \frac{S_s}{1 + \cosh((T-s)\kappa) } ds\right ]\]

with trading rate $v^*_t = -\dot{X}^*_t$, such that:
    \[v^*_t = X^*_t \kappa \coth(\kappa(T-t)) + \frac{\lambda S_t}{2\kappa} \tanh \left( \frac{\kappa (T-t)}{2}\right)\]

In addition, the closed-form value function is given by:
\[\Gamma^* (\tau, X ,S) = \kappa X^2 \coth (\tau\kappa) + \frac{\lambda X S}{\kappa} \tanh \left (\frac{\tau \kappa}{2} \right) - \frac{\lambda^2 S^2 e^{\sigma^2 \tau}}{4\kappa^2} \int^\tau_0 \left [\tanh \left(\frac{u \kappa}{2} \right ) \right ]^2 e^{-\sigma^2 u} du \]

\begin{corollary}[Zero Risk Penalty Value Function Symmetry Property] \label{col:OE-sym-0}
    For the value function defined in Proposition \ref{prop:OE-HJB-zero} the following symmetry holds true:
    \begin{align}
        \Gamma(\tau, X) = \Gamma(\tau, -X)
    \end{align}
\end{corollary}
\begin{proof}
\begin{align*}
    \Gamma(\tau, -X) &= \inf_{v \in \mathbb{R}} \mathbb{E}\left [  \int^\tau_0 v^2_t + \kappa^2 (-X_t)^2 dt\right] \\ 
    &= \inf_{v \in \mathbb{R}} \mathbb{E}\left [  \int^\tau_0 v^2_t + \kappa^2 X_t^2 dt\right] \\
    &= \Gamma(\tau, X)    
\end{align*}
\end{proof}

\begin{corollary}[Positive Risk Penalty Value Function Symmetry Property]\label{col:OE-sym-lam}
    For the value function defined in Proposition \ref{prop:OE-HJB-pos} the following symmetry holds true:
    \begin{align}
        \Gamma(\tau, X, S) = \Gamma(\tau, -X, -S)
    \end{align}
\end{corollary}
\begin{proof}
    \begin{align*}
        \Gamma(\tau, -X, -S) &= \inf_{v \in \mathbb{R}} \mathbb{E}\left [  \int^\tau_0 v^2_t + \kappa^2 (-X_t)^2  + \lambda (-X_t) (-S_t)dt\right] \\ 
    &= \inf_{v \in \mathbb{R}} \mathbb{E}\left [  \int^\tau_0 v^2_t + \kappa^2 X_t^2  + \lambda X_t S_t dt\right] \\
    &= \Gamma(\tau, X, S)
    \end{align*}
\end{proof}

\section{Core MT-PINN framework} \label{Appendix B}

\subsection{Generic Architecture} \label{Appendix B1}
The objective is to approximate the scalar value function $\Gamma(\tau, \xi)$ with a smooth neural network $\hat{\Gamma}(\tau, \xi; \theta)$ that supports accurate derivatives for HJB residuals. The PINNs used in this report, including MT-PINN, were developed in JAX/FLAX, using a single-head Multilayer Perceptron (MLP) with inputs $z=(\tau, \xi) \in \mathbb{R}^{1+d}$, with $\tau \in [0, T]$ (time-to-maturity) and $\xi \in \mathbb{R}^d$ (state), and with tanh smooth activation. Normalization layers and stochastic regularizers are avoided to preserve stable derivatives. Automatic differentiation was used to obtain $\partial_\tau \hat{\Gamma}, \nabla_\xi \hat{\Gamma}$, and second-order derivatives of the Hessian, where required, for computing the HJB residuals. Although the optimal control is not specifically parameterized, it is recovered analytically via these derivatives. MT-PINNs use the formulated optimal control derived from the HJB equation to then unroll the system dynamics along sampled trajectories and define a trajectory-based loss (viz., terminal zero-state and running-cost penalties) computed via backpropagation through time (BPTT), thereby coupling the HJB residuals with realized control performance. The intention of this approach is that the trajectory-based loss not only naturally enforces the terminal zero-state condition, but also ensures that the value function along the optimal path is accurate.

\subsection{MT-PINN training mechanics (BPTT) \& complexity} \label{Appendix B2}
At each training step, trajectories, via forward Euler discretization, are simulated. At each time step the network is queried to produce the control. After reaching the terminal time, the terminal trajectory loss is computed and then Backpropagation Through Time (BPTT) is used to propagate sensitivities backward through all time steps to obtain the gradient with respect to the network parameters of that trajectory loss. Then an Adam optimizer performs one parameter update using the sum of gradients from all loss terms.

For a single trajectory, consider the loss:

\[\mathcal{L}(\theta) = \sum^{N-1}_{k=0} l_k(x_k) + l_N(x_N), \]
where $l_k$ are all the running penalties and $l_N$ is the terminal penalty. In many applications, a terminal-only penalty is used (i.e., $l_k \equiv 0$ for $k < N$). Nonetheless, the derivation below covers both cases.

Let $\theta$ denote all trainable network parameters and $x_k \in \mathbb{R}^n$ denote the state at step $k=0,..,N$. Let $z_k$ be the per-step exogenous inputs. A differentiable step map $f_\theta  : \mathbb{R}^n \times \mathcal{U} \rightarrow \mathbb{R}^n$ advances the state by one time step: 

\begin{equation*}
    x_{k+1} = f_\theta(x_k, z_k), 
\end{equation*}
For an explicit forward Euler discretization of continuous dynamics $\dot{x} = g_\theta(x, z)$, we iterate the step map to obtain $x_1, ...., x_N$ and evaluate $\mathcal{L}(\theta)$ as follows:

\begin{equation*}
     x_{k+1} =  x_k +  g_\theta(x_k,z_k)\Delta t,
\end{equation*}
assuming that $f_\theta$ is differentiable in both $x$ and $\theta$. 

Then define the sensitivity/adjoint:
\[\lambda_k := \frac{\partial \mathcal{L}}{\partial x_k} \in \mathbb{R}^n,\]
and define the local Jacobians

\begin{equation*}
    A_k  := \frac{\partial f_\theta}{\partial x_k}(x_k,z_k) \in \mathbb{R}^{n \times n}, \ \ \ \ B_k := \frac{\partial f_\theta}{\partial \theta}(x_k, z_k) \in \mathbb{R}^{n \times |\theta|}
\end{equation*}
BPTT then follows as:

\begin{align*}
    \text{(initialization)} \ \ \ \lambda_N = \frac{\partial l_N}{\partial x_N}(x_N),\\
    \text{(backward recursion)} \ \ \ \lambda_k = \frac{\partial l_k}{\partial x_k}(x_k) + A^\intercal_k \lambda_{k+1}, \ \ k=N-1,...,0, \\ 
    \text{(parameter gradient)} \ \ \ \nabla_\theta \mathcal{L} = \sum^{N-1}_{k=0} \left ( B^\intercal_k \lambda_{k+1} + \frac{\partial l_k}{\partial \theta} \right ) + \frac{\partial l_N}{\partial \theta}
\end{align*}
In the case of terminal-only penalties, more specifically when $l_k \equiv 0, \ \text{for}  \ k <N \Rightarrow \partial l_k/\partial x_k = 0$, this can be simplified to:

\begin{align*}
    \text{(backward recursion)} \ \ \ \lambda_k = A^\intercal_k \lambda_{k+1}, \ \ k=N-1,...,0, \\ 
    \text{(parameter gradient)} \ \ \ \nabla_\theta \mathcal{L} = \sum^{N-1}_{k=0} \left (B^\intercal_k \lambda_{k+1} + \frac{\partial l_k}{\partial \theta} \right ),
\end{align*}
Practically, modern frameworks (e.g., JAX) implement this via reverse-mode autodiff, by applying vector-Jacobian products on the step function at each step to obtain updated sensitivity and parameter gradients.

\paragraph{Complexity.}
Let $N_\text{dt}$ be the Euler steps per simulation, $B$ be the number of trajectories in a batch (i.e., number of initial states \(\times\) horizons sampled), and $C_{f_\theta}$ represents the cost of evaluating the network $f_\theta$. 

\textbf{Time per epoch:} \(\mathcal{O}(N_\text{dt} \times B \times C_{f_\theta})\). 

\textbf{Space per epoch:} \(\mathcal{O}(N_\text{dt} \times B \times A)\), where \(A\) denotes the per-step activation memory per network evaluation saved for backpropagation.

Note that these complexities are equivalent to \(\mathcal{O}(J \times N_\text{dt} \times P)\) used in the main body under \(B= P \times J\). Furthermore, this new penalty is also amenable to parallelism as simulated trajectories are independent and vectorized.

\paragraph{Compute resources} Experiments were run on 1x v6e-1 TPU and constructed using JAX/FLAX. Typical runtime ranged from 1 minute to 4 minutes.

\subsection{Generic Loss Template} \label{Appendix B3}

Let $\mathcal{C}$ denote the set of constraint/boundary conditions that are application-specific. The total loss for MT-PINNs can be then expressed as:
$$\mathcal{L}_\text{total} = w_\text{PDE} \mathcal{L}_\text{PDE} + w_\text{traj} \mathcal{L}_\text{traj} + \sum_{\ell \in \mathcal{C}} w_\ell \mathcal{L}_\ell,$$
$$\mathcal{L}_\text{PDE} = \mathbb{E}_{z \sim \mathcal{D}_\text{PDE}} \left [ \mathcal{R}_\text{HJB} (z; \theta)^2\right ], \ \ \mathcal{L}_\text{traj} = \mathbb{E}_{\zeta \sim \mathcal{D}_\text{traj}} \left [ \psi(x_T (\zeta ; \theta) ) ] \right ], \ \ \psi(x_T) = \begin{cases} |x_T|,& |x_T| \le 1, \\ x^2_T,& |x_T|  > 1. \end{cases}$$
where 
\begin{itemize}
    \item $\mathcal{D}_\text{PDE}$: is the sampling distribution of PDE collocation points over $z$.
    \item  $\mathcal{D}_\text{traj}$: is its sampling distribution.
    \item $\mathcal{R}_\text{HJB}(z;\theta)$: is the HJB/PDE residual at $z$ using the network.
    \item $\zeta$: is a trajectory sample tuple.
    \item $x_T(\zeta; \theta)$: is the state at terminal time from rolling out dynamics under the control implied by the network starting at $\zeta$.
    \item $\psi(x_T)$: is the terminal penalty, defined as a piecewise function to enforce zero state/inventory at maturity.
    \item $w_\text{PDE}, w_\text{traj}, \{ w_\ell \}$ are non-negative.
\end{itemize}

\begin{figure}[hbt!]
    \centering
    \fbox{%
      \includegraphics[width=\linewidth]{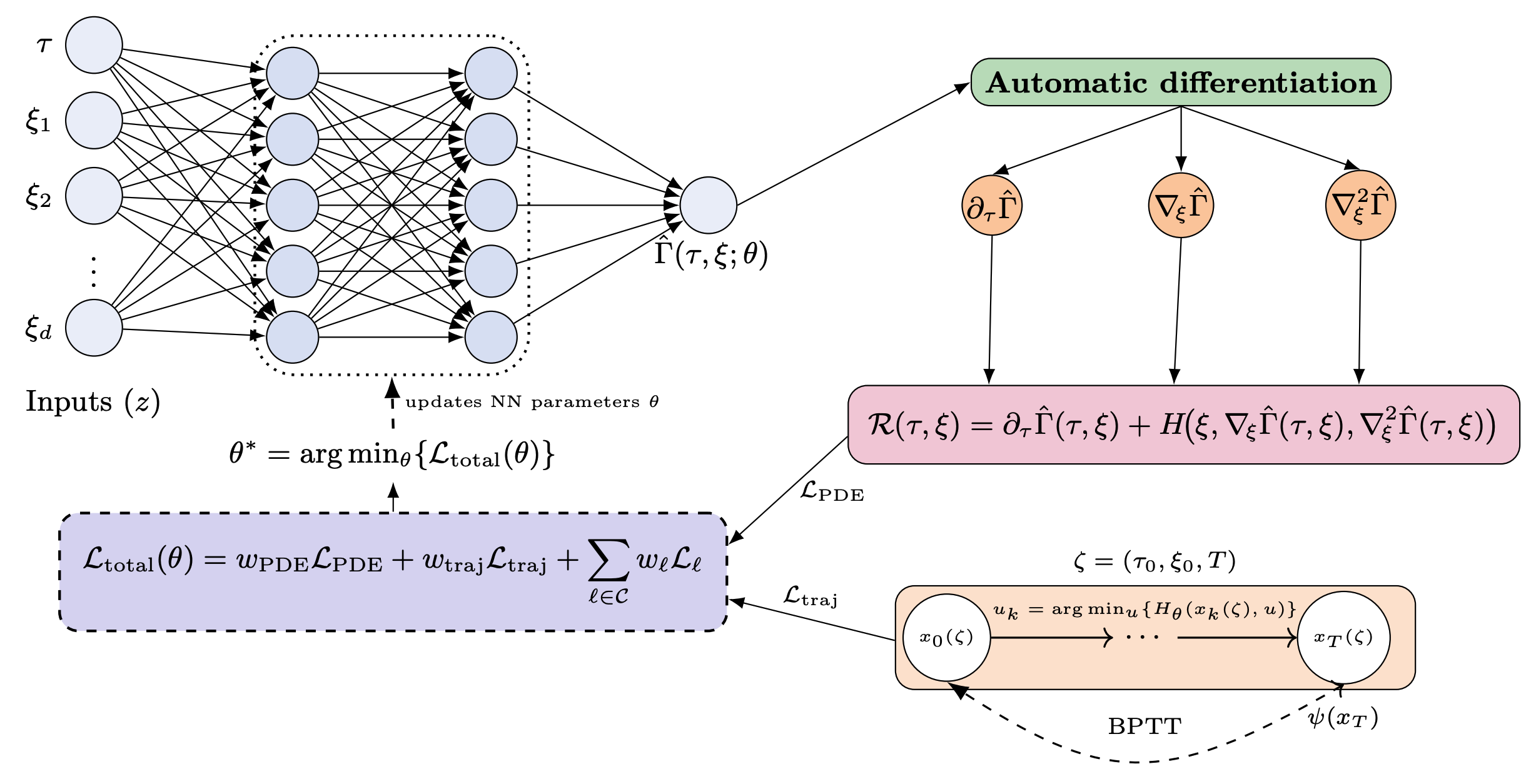}%
      }
    \caption{Core MT-PINN framework diagram. A neural network $\hat{\Gamma}(\tau, \xi;\theta)$ outputs an approximation of the value function. Automatic differentiation then provides the derivatives for the PDE. Using this, the PDE residual $\mathcal{R}(\tau, \xi)$ and the controls $u_k$ are calculated. The PDE residual forms the PDE loss function $\mathcal{L}_\text{PDE}$ and the controls form the trajectory rollout penalty $\mathcal{L}_\text{traj}$ via BPTT. Both, contribute to the total loss $\mathcal{L}_\text{total}(\theta)$, which updates the neural network parameters $\theta$ via gradient descent.}
    \label{fig:MT_diagram}
\end{figure}

See Figure \ref{fig:MT_diagram} for an illustration of the core MT-PINN framework.

\subsection{Core MT-PINN framework diagram} \label{Appendix B4}
\FloatBarrier

\section{Full MT-PINN Losses \& Weighting (DWA)} \label{Appendix C}

\subsection{Full Loss (by regime)} \label{Appendix C.1}
The total loss function for MT-PINN is expressed as a piecewise function depending on whether the risk-aversion is zero or positive:

\[\mathcal{L}_\text{total} = 
\begin{cases} 
w_\text{PDE} \mathcal{L}^{(0)}_\text{PDE} + w_\text{traj} \mathcal{L}^{(0)}_\text{traj} + w_\text{IC} \mathcal{L}_\text{IC}^{(0)} + w_\text{sym} \mathcal{L}_\text{sym}^{(0)}, & \lambda = 0, \\
w_\text{PDE} \mathcal{L}_\text{PDE}^{(\lambda)} + w_\text{traj} \mathcal{L}^{(\lambda)}_\text{traj} + w_\text{IC} \mathcal{L}_\text{IC}^{(\lambda)} + w_\text{sym} \mathcal{L}_\text{sym}^{(\lambda)} + w_\text{0-term} \mathcal{L}_\text{0-term}, & \lambda > 0.
\end{cases}\]

\paragraph{PDE loss}
Let $\Gamma_\theta$ denote the value function from the network, and its partial derivatives are $\Gamma_\tau = \frac{\partial \Gamma_\theta}{\partial \tau}$, $\Gamma_X = \frac{\partial \Gamma_\theta}{\partial X}$, $\Gamma_S = \frac{\partial \Gamma_\theta}{\partial S}$, and $\Gamma_{SS} = \frac{\partial^2 \Gamma_\theta}{\partial S^2}$. The PDE residual uses the HJB form according to the regime:
\begin{align*}
    \text{For}  \ \lambda = 0:& \ \ \mathcal{R}^{(0)}_\text{HJB} = \Gamma_\tau - \kappa^2X^2 + \frac{1}{4} \Gamma_X^2,  \ \ \ \text{(Proposition \ref{prop:OE-HJB-zero})} \\ 
    \text{For } \lambda > 0:& \ \ \mathcal{R}^{(\lambda)}_\text{HJB} = \Gamma_\tau - \frac{1}{2} \sigma^2 S^2 \Gamma_{SS} - \kappa^2X^2  - \lambda S X + \frac{1}{4}\Gamma^2_X. \ \ \ \text{(Proposition \ref{prop:OE-HJB-pos})}
\end{align*}
$\mathcal{L}^{(\cdot)}_\text{PDE}$ is then defined as:

\begin{align*}
    \mathcal{L}^{(0)}_\text{PDE} & =  \frac{1}{N^{(0)}_\text{PDE}} \sum^{N^{(0)}_\text{PDE}}_{i=1} \left [ \mathcal{R}^{(0)}_\text{HJB}(\tau_i, X_i)\right]^2, \ \ \ (\tau_i, X_i) \sim \mathcal{D}^{(0)}_\text{PDE},\\
    \mathcal{L}^{(\lambda)}_\text{PDE} &= \frac{1}{N^{(\lambda)}_\text{PDE}} \sum^{N^{(\lambda)}_\text{PDE}}_{i=1} \left [ \mathcal{R}^{(\lambda)}_\text{HJB}(\tau_i, X_i, S_i)\right]^2, \ \ \ (\tau_i, X_i, S_i) \sim \mathcal{D}^{(\lambda)}_\text{PDE},
\end{align*}
\paragraph{Inventory-zero loss (internal condition at \(X = 0\).}
The loss of the internal condition at \(X=0\), defined by (\ref{eq:OE-BC-0}) for the risk-neutral regime and (\ref{eq:OE-BC-lam}) for the risk-averse regime, is as follows:

\begin{align*}
    \mathcal{L}^{(0)}_{IC}  &\approx \frac{1}{N^{(0)}_{IC}} \sum^{N^{(0)}_{IC}}_{i=1} \Gamma_\theta(\tau_i, 0)^2, \ \ \tau_i \sim \mathcal{D}^{(0)}_{IC} \\
    \mathcal{L}^{(\lambda)}_{IC}  &\approx \frac{1}{N^{(\lambda)}_{IC}} \sum^{N^{(\lambda)}_{IC}}_{i=1} \left [ \max \{ \Gamma_\theta(\tau_i, 0, S_i), 0\} \right ]^2, \ \ \tau_i,S_i \sim \mathcal{D}^{(\lambda)}_{IC}. \\
\end{align*}
\paragraph{Symmetry loss.}
The symmetric condition loss encodes Corollary \ref{col:OE-sym-0} for the risk-neutral regime and Corollary  \ref{col:OE-sym-lam} for the risk-averse regime as:

\begin{align*}
    \mathcal{L}^{(0)}_\text{sym} &\approx \frac{1}{N^{(0)}_\text{sym}} \sum^{N^{(0)}_\text{sym}}_{i=1} \left [ \Gamma_\theta (\tau_i, X_i) - \Gamma_\theta (\tau_i, -X_i) \right]^2, \ \ \tau_i, X_i \sim \mathcal{D}^{(0)}_\text{PDE}, \\
    \mathcal{L}^{(\lambda)}_\text{sym} &\approx \frac{1}{N^{(\lambda)}_\text{sym}} \sum^{N^{(\lambda)}_\text{sym}}_{i=1} \left [ \Gamma_\theta (\tau_i, X_i, S_i) - \Gamma_\theta (\tau_i, -X_i, -S_i) \right]^2, \ \ \tau_i, X_i, S_i \sim \mathcal{D}^{(\lambda)}_\text{PDE}.
\end{align*}

\paragraph{Terminal zero-inventory loss.}
Due to the internal condition imposed on the risk-averse regime of (\ref{eq:OE-BC-lam}), a separate loss term was created to ensure the network learns that $\Gamma(0,0,S) = 0$ defined as:

\[\mathcal{L}_\text{0-term} \approx \frac{1}{N_\text{term}} \sum^{N_\text{term}}_{i=1} \Gamma(\tau = 0, X = 0, S_i)^2, \ \ \ S_i \sim \mathcal{D}_\text{term}.\]
Note that the risk-neutral regime already explicitly imposes this constraint on all ($\tau,S$) for $X=0$.

\paragraph{Sampler definitions.}
\(\mathcal{D}\) denotes the uniform sampling distribution for each collocation domain. Experiment-specific grids and counts (i.e., \(N_{(\cdot)}\), horizon sets, and initial state grids) are provided in the respective experiment appendices.

\subsection{DWA-style weight rebalancing} \label{Appendix C.2}
We adapt task weights with a dynamic weight average (DWA)-style scheme \citep{liu2019end}. Instead of using DWA's raw ratio \(L_\mathcal{T}(t-1)/L_\mathcal{T}(t-2)\) and softmax, we use a smoothed exponential moving average (EMA) of each loss and update weights from its relative scale. Concretely, we make the following lightweight stabilizations:
\begin{itemize}
    \item EMA smoothing of losses, relative to initial scaling
    \item Geometric mean centering across task 
    \item Gentle inverse power-law update instead of softmax and clip weights to \([w_\text{min}, w_\text{max}]\)
    \item Mean-one normalization of weights across active tasks each update
    \item Optional freezing for tasks that become tiny over several updates
    \item Scheduled updates (i.e., updating weights every \(\Delta\) epochs)
\end{itemize}
Hyperparameters are listed within the respective experiment appendices.

\section{Synthetic benchmark: complete experiment specification} \label{Appendix D}

The MT-PINN used for this experiment is outlined in Appendix \ref{Appendix C.1}. As a replacement for the trajectory loss, which aims to naturally fulfill the terminal condition stated in equations (\ref{OE-term-0}) and (\ref{OE-term-lam}), both PINN with $\lambda$-curriculum and vanilla PINN have a loss term that sets an arbitrarily large penalty for leftover inventory to fulfill the terminal condition. This loss term is formulated as:

\begin{align*}
    \mathcal{L}^{(0)}_\text{term}&= \frac{1}{N^{(0)}_\text{term}} \sum^N_{i=1} \left (\Gamma_\theta (0,X_i)-cX^2_i \right)^2, \ \ X_i \sim \mathcal{D}^{(0)}_\text{term}, \\
 \mathcal{L}^{(\lambda)}_\text{term}&= \frac{1}{N^{(\lambda)}_\text{term}} \sum^N_{i=1} \left (\Gamma_\theta (0,X_i,S_i)-cX^2_i \right)^2, \ \ X_i, S_i \sim \mathcal{D}^{(\lambda)}_\text{term},
\end{align*}
where $c > 0$  is the penalty strength. Otherwise the loss terms across the PINN variant models are the same:
\[\mathcal{L}_\text{total} = 
\begin{cases} 
w_\text{PDE} \mathcal{L}^{(0)}_\text{PDE}  + w_\text{IC} \mathcal{L}_\text{IC}^{(0)} + w_\text{sym} \mathcal{L}_\text{sym}^{(0)} + w_\text{term} \mathcal{L}^{(0)}_\text{term}, & \lambda = 0, \\
w_\text{PDE} \mathcal{L}_\text{PDE}^{(\lambda)} + w_\text{IC} \mathcal{L}_\text{IC}^{(\lambda)} + w_\text{sym} \mathcal{L}_\text{sym}^{(\lambda)} + w_\text{term} \mathcal{L}_\text{term}^{(\lambda)}, & \lambda > 0.
\end{cases}\]

For the synthetic benchmark experiment, three values of risk-aversion were evaluated: $\lambda^* \in \{0, 0.05, 0.1\}$. For $\lambda = 0$ all methods are trained for 30k epochs. For $\lambda > 0 $, vanilla PINN was trained for 55k epochs, while the curriculum-based models used a two-phase schedule totaling 55k epochs (30k at $\lambda = 0$, then 5 curriculum stages of 5k epochs each up to the target $\lambda^*$).

\begin{figure}
    \centering
    \includegraphics[width=\linewidth]{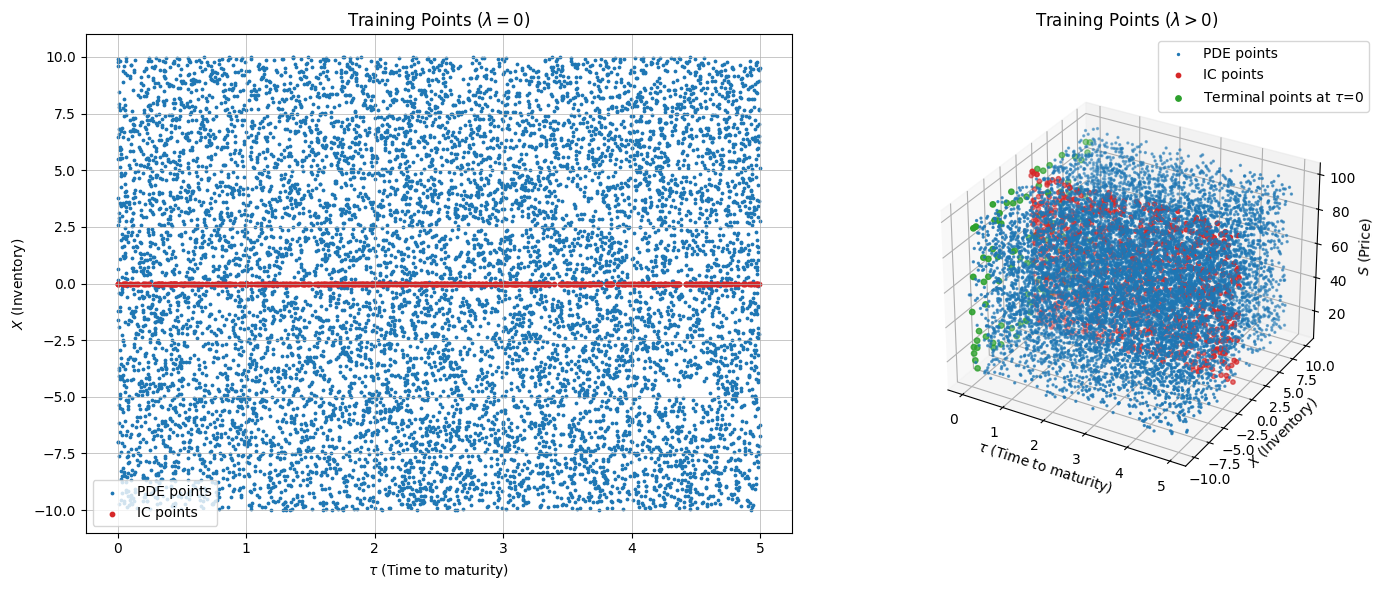}
    \caption{Sampling/collocation points}
    \label{fig:samp-dist}
\end{figure}

\textbf{Common setting (unless stated otherwise).}
Time horizon $T=5.0$, state ranges $X \in [-10,10], S \in [10, 100]$ (the $S$ range only used when $\lambda > 0$). The PDE collocation points were $N_{PDE} = 30,000$ and internal conditions points were $N_{IC} = 5000$ (see \ref{fig:samp-dist}). Model parameters: $\kappa = 0.1$ (risk-adjusted inventory penalty), $\sigma = 0.1$ (price volatility). Optimization uses AdamW with learning rate $5 \times 10^{-4}$. Loss weights are adapted with DWA-style scheme as briefly described in Appendix \ref{Appendix C.2} with weights updated every \(\Delta = 1000\) epochs, weight clipping \([0.1, 2.0]\), EMA smoothing factor \(\beta = 0.95\) , update strength \(\alpha = 0.3\), and freeze tolerance of \(10^{-4} \). Initial weights were set to \([w_\text{PDE} = 1.0, w_\text{traj} = 1.0, w_\text{IC} = 0.1, w_\text{sym}=0.5, w_\text{0-term}=0.5, w_\text{term}=1.0]\).

\textbf{Vanilla PINN (single-stage).}
MLP width of (500, 500, 500). For $\lambda = 0$: 30k epochs. For $\lambda \in \{0.05,0.10 \}$: 55k epochs with the same sampler sizes and network input dimensions were set by the regime. Terminal points were $N_\text{term} = 5000$.

\textbf{PINN + $\lambda$-Curriculum (warm-start 1D state to 2D state).}
Phase A (1D, $\lambda = 0$): MLP width of (500, 500, 500), two input dimensions, and run for 30k epochs.
Phase B (2D, $\lambda > 0$): warm start the trained 1D parameters to 2D then train 5 curriculum stages with $\alpha \in (0.25, 0.5, 0.75, 0.9, 1.0)$. Each stage runs 5k epochs, with $\lambda_\alpha = \alpha \cdot \lambda^*$. Terminal points were $N_\text{term} = 5000$.

\textbf{MT-PINN + \(\lambda\)-Curriculum (warm-start 1D state to 2D state).}
Phase A (1D, $\lambda = 0$): MLP width of (32, 32, 32), two input dimensions, and run for 30k epochs.
Phase B (2D, $\lambda > 0$): warm start the trained 1D parameters to 2D then train 5 curriculum stages with $\alpha \in (0.25, 0.5, 0.75, 0.9, 1.0)$. Each stage runs 5k epochs, with $\lambda_\alpha = \alpha \cdot \lambda^*$. Multi-trajectory batch samples \(P=820\) (\(n_X = 41\) and where applicable \(n_S = 20\)) and roll out horizons of \(\{T /50, T /10, T /5, 2T /5, 3T /5, 4T /5, T \}\) with \(N_\text{dt} = 200\) Euler steps. Terminal points were $N_\text{term} = 2000$.

\section{Real-market SPY backtest: data, preprocessing, metrics, \& experiment specification}\label{Appendix E}
\subsection{Data \& preprocessing}\label{Appendix E.1}
Intraday market data for SPY was obtained via Databento, sourced from the Nasdaq TotalView-ITCH feed. Data was used under Databento's academic/historical data provisions, subject to Nasdaq TotalView-ITCH license/ToS. We cannot redistribute the raw data.

The trading day is split into three 2-hour fixed intraday windows on 5-second intervals, and repeated across seven days (Feb 10 - Feb 19, 2025, excluding weekends). The windows are:
\begin{itemize}
    \item Window 1 (W1): 09:45 - 11:45 ET
    \item     Window 2 (W2): 11:45 - 13:45 ET
    \item Window 3 (W3): 13:45 - 15:45 ET

\end{itemize}   

This intraday segmentation of SPY mid-price is illustrated in Figure \ref{fig:spy_intraday_windows}. Note that these windows exclude the first 15 minutes of market open and the last 15 minutes of market close. This was done to avoid the noisy and elevated volatilities typically observed at those times \citep{INCI201779}. 
\FloatBarrier

\begin{figure}
    \centering
    \includegraphics[width=\textwidth]{Market/SPY_prices.png}
    \caption{
        Intraday segmentation of SPY mid-price series into fixed windows. 
        Each trading day is partitioned into three non-overlapping 2-hour intervals: 
        Window 1 (W1) from 09:45 to 11:45 ET, 
        Window 2 (W2) from 11:45 to 13:45 ET, and 
        Window 3 (W3) from 13:45 to 15:45 ET. 
        Vertical dashed lines mark the boundaries between windows, while solid black vertical lines indicate the start of each new trading day.
    }
    \label{fig:spy_intraday_windows}
\end{figure}
\subsection{Metrics}\label{Appendix E.2}

The main criteria for comparing these models were their exposure and cost trade-off. Exposure measures how much inventory the strategy was carrying on average for a given window. This was calculated using

\[\text{Exposure}(W_j) = \frac{1}{N}\sum^{N-1}_{k=0} \left ( \chi^{(j)}_k\right )^2,\]
where $W_j$ is the $j^\text{th}$ trading window, $N$ is the number of discretized equal steps, $\chi_k$ is the normalized inventory at step $k$. Cost is measured as the implementation shortfall relative to liquidating the entire initial position $X_0$ at the initial mid-price $S_0$. In practice, this measures the cost associated with using the execution model for liquidation versus dumping the entire position immediately at the current mid-price. Cost was measured in basis points  (1 bps = 0.01\%) and computed using 

\[ \text{Cost}(W_j) = \frac{S^{(j)}_0 X_0 - \sum^{N-1}_{k=0} q^{(j)}_k S^{(j)}_k}{S^{(j)}_0 X_0} \times 10^4 \ \ \text{bps},\]
where $q_k$ denotes the trade at step $k$. The exposure and cost across all the trading windows are then aggregated to report the mean and standard deviation of the exposure and cost using NumPy built-in functions (see Table \ref{tab:cost_variance} for results).

\subsection{Experiment specification}\label{Appendix E.3}

Part of applying these models into real-market data meant adjusting the model parameters so that they accurately match the data. 

\textbf{Time Normalization.}
Time horizon was expressed by trading day units, where elapsed time is normalized by the effective length of a trading day on the U.S. equity markets (6.5 hours, from 9:30 a.m. to 4:00 p.m. EST). Specifically, the two-hour backtest horizon corresponded to

\[T_\text{days} = \frac{2}{6.5} \approx 0.308,\]
which represents roughly 31\% of the trading day.

\textbf{Inventory Normalization.}
Inventory was normalized by the initial position, so that the inventory lies in the interval $[-1,1]$, and represents the fraction of the initial inventory:

\[\chi_t = \frac{x_t}{X_0},\]
where $\chi_t = 1$ corresponds to holding the full initial inventory, $\chi_t =0$ to fully liquidated inventory, and $\chi_t = -1$ corresponds to over-liquidation (shorting) of equal size of the initial inventory.

\textbf{Stock Price Range.}
Over the full 21 time windows, the stock price range was: $\$599.60 - \$613.20$. Therefore, the $S$ range used for MT-PINN was $590-620$, leaving a small buffer of roughly 1\% each side.

\textbf{Volatility Estimation.}
To estimate volatility ($\sigma$) of the assumed unaffected price process $S_t$, the realized volatility for each intraday execution window was computed and then averaged. 
Given a set of stock mid-prices $\{S_{t_i}\}^N_{i=1}$ in a given window of length $\Delta T_\text{W}$, the log-returns can be calculated as
\[r_i = \ln \left ( \frac{S_{t_i}}{S_{t_{i-1}}} \right ), \ \ \ i=1,...,N.\]
Hence the realized volatility for a given window is

\[\hat{\sigma}^2_\text{W} = \frac{1}{N-1} \sum^N_{i=1} (r_i - \bar{r})^2,\]
where $\bar{r}$ denotes the sample mean of returns. For window length $\Delta T_W$, realized volatility was rescaled into trading-day units as 

\[\hat{\sigma}_\text{W, daily} =\frac{\hat{\sigma}_\text{W}}{\sqrt{\Delta T_\text{W}}}.\]
Finally, averaging across all execution windows yields the representative daily volatility parameter for the simulation:

\[\sigma = \frac{1}{J} \sum^J_{j=1} \hat{\sigma}^{(j)}_\text{W, daily},\]
where $J$ represents the number of execution windows. From the SPY market data, the calculated volatility was:

\[\sigma \approx 0.0038 \  \ \ \ (\approx 6 \%\text{ annualized}).\]

\textbf{MT-PINN (Phase A: 1D, Phase B: 2D).}
MT-PINN with width (32,32,32), scalar output, and tanh activations. Time horizon $T \approx 0.308$, with horizon grid $T_j  \in \{T/20, T/10, T/4, T/2, 3T/4, T \},$ $Ndt_\text{traj} = 50$ Euler steps and $n_X = 200$ inventory grid points. Inventory range set to $[-1, 1]$ and $\kappa =0.2$. This experiment uses the same hyperparameters for the weights as outlined in Appendix \ref{Appendix D}, with the exception to weighting clipping \([0.1, 5.0]\) and initial trajectory loss weighting \(w_\text{traj} = 5.0\).

Phase A: uses $(\tau, X)$ inputs, 20k training epochs with 30k collocation points, and risk-neutral regime $\lambda =0$.
Phase B: uses $(\tau, X, S)$ with price $S \in [590,  620]$, discretized with $n_S = 41$ for multi-trajectory loss, and trained on 20k collocation points for 5k epochs. Volatility set to $\sigma = 0.0038$. Risk aversion parameter introduced through curriculum stages $\alpha \in \{ 0.25, 0.5, 0.75, 1.0\}$ for $\lambda^*=[0.05, 0.1]$, with $\lambda_\alpha = \alpha\cdot \lambda^*$. Sample sizes: $N_{IC} = 2000, N_\text{0-term} = 200$. 

\section{Further Synthetic Benchmark \& SPY Real-Market Backtest Results} \label{Appendix F}
\subsection{Synthetic Benchmark Results}

\FloatBarrier

\begin{figure}
  \centering
  \begin{subfigure}[t]{0.45\linewidth}
    \centering
    \includegraphics[width=\linewidth]{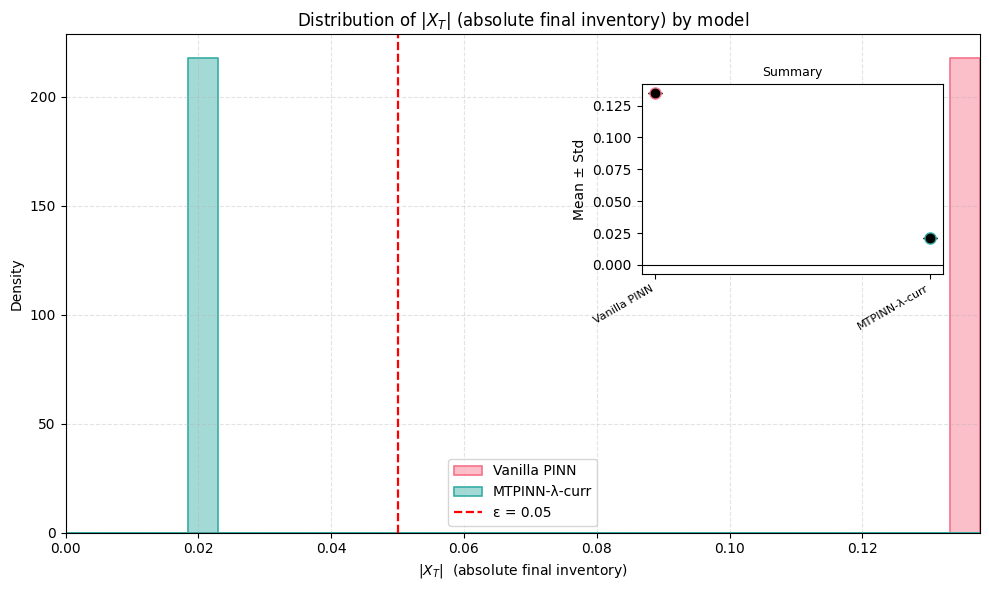}%
    \caption{$\lambda=0$}
  \end{subfigure}\hfill
  \begin{subfigure}[t]{0.45\linewidth}
    \centering
    \includegraphics[width=\linewidth]{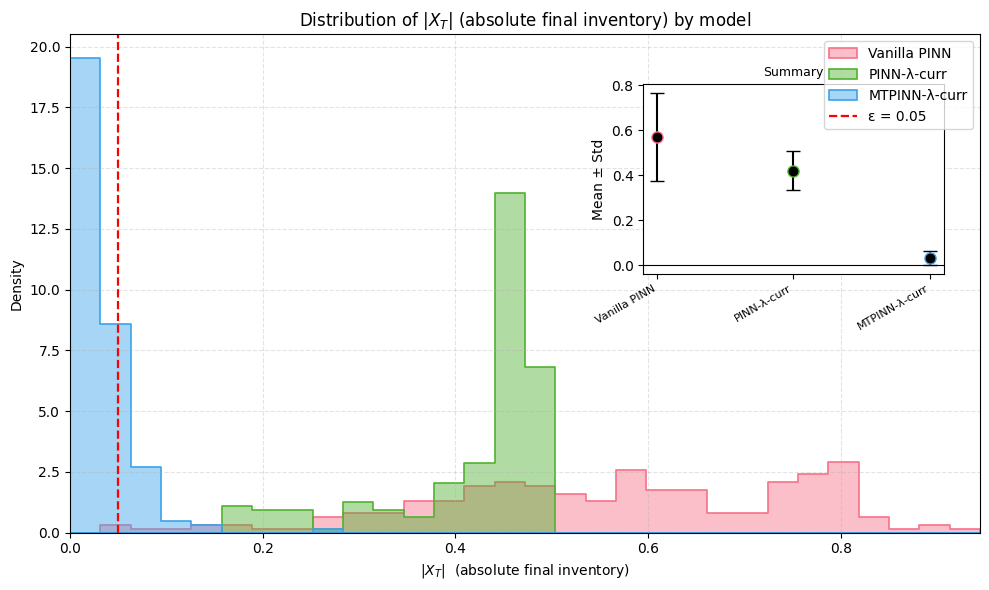}%
    \caption{$\lambda=0.05$}
  \end{subfigure}\hfill
  \begin{subfigure}[t]{0.45\linewidth}
    \centering
    \includegraphics[width=\linewidth]{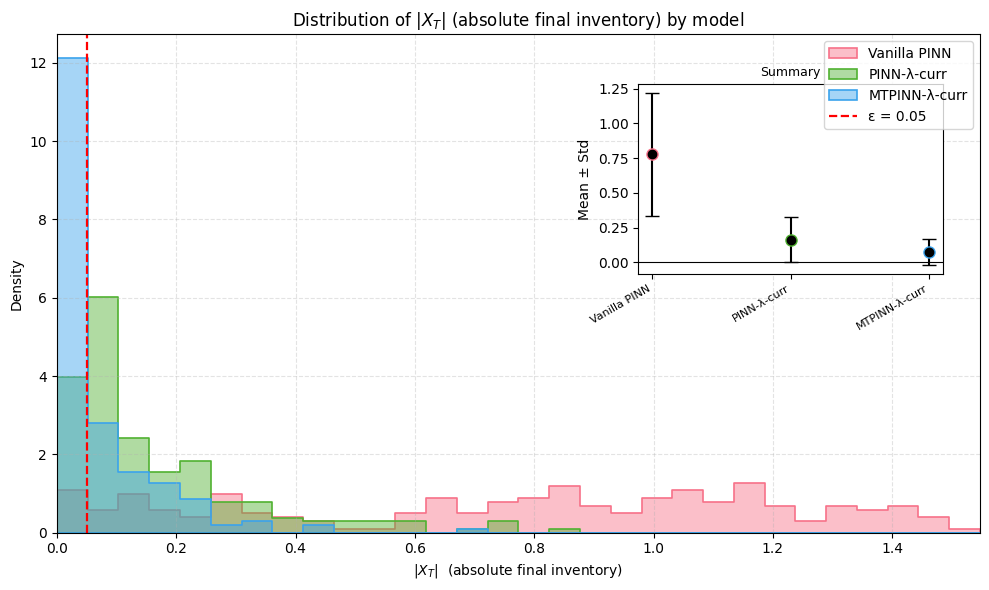}%
    \caption{$\lambda=0.10$}
  \end{subfigure}

  \caption{Distribution of \(\lvert X_T\rvert\) (absolute final inventory) by model for each risk aversion \(\lambda\).
  Each panel shows the histogram across rollouts for the three models
  (Vanilla PINN, PINN-\(\lambda\)-curr, MT-PINN-\(\lambda\)-curr; PINN-\(\lambda\)-curr omitted for $\lambda=0$) with the tolerance line at \(\epsilon=0.05\) and the inset reporting mean\(\,\pm\)std.
  }
  \label{fig:final-inventory-dists}
\end{figure}

\begin{figure}
  \centering
  \begin{subfigure}[t]{0.5\linewidth}
    \centering
    \includegraphics[width=6cm]{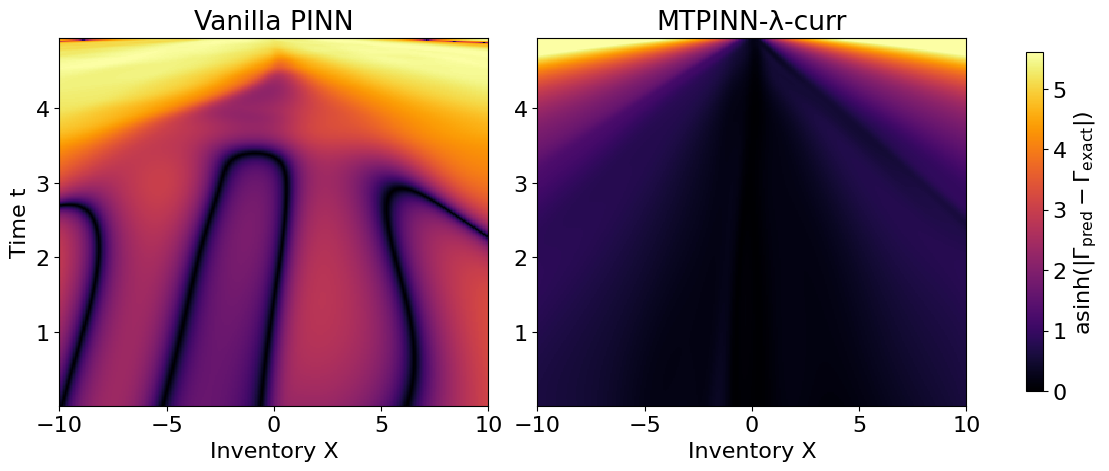}
    \caption{$\lambda=0$}
  \end{subfigure}\hfill
  \begin{subfigure}[t]{0.5\linewidth}
    \centering
    \includegraphics[width=\linewidth]{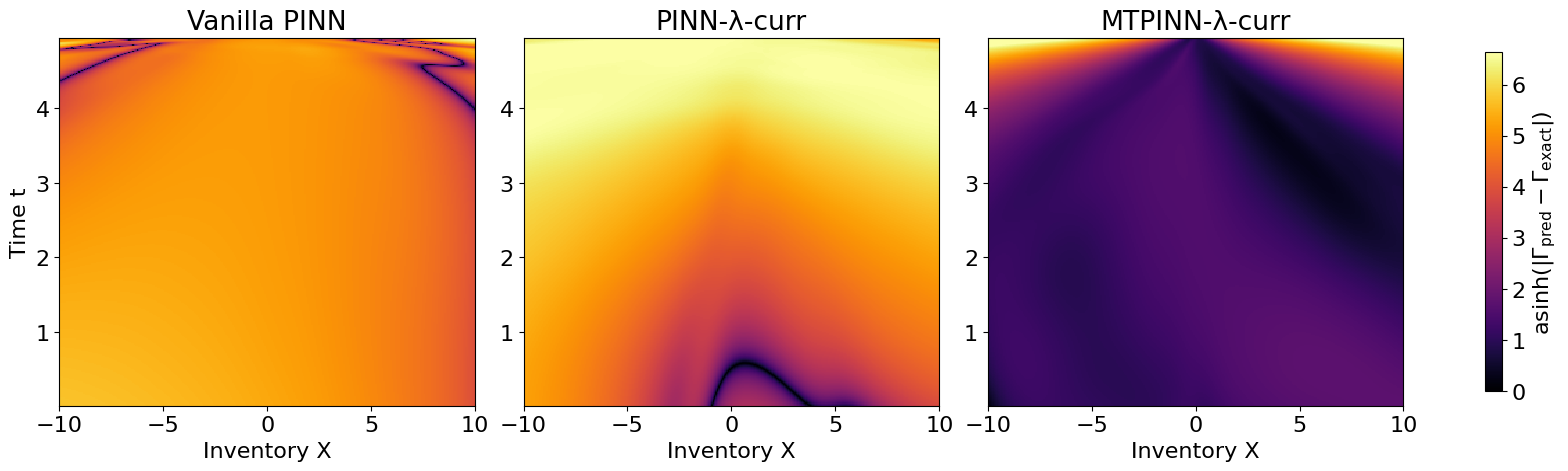}
    \caption{$\lambda=0.05$}
  \end{subfigure}\hfill
  \begin{subfigure}[t]{0.5\linewidth}
    \centering
    \includegraphics[width=\linewidth]{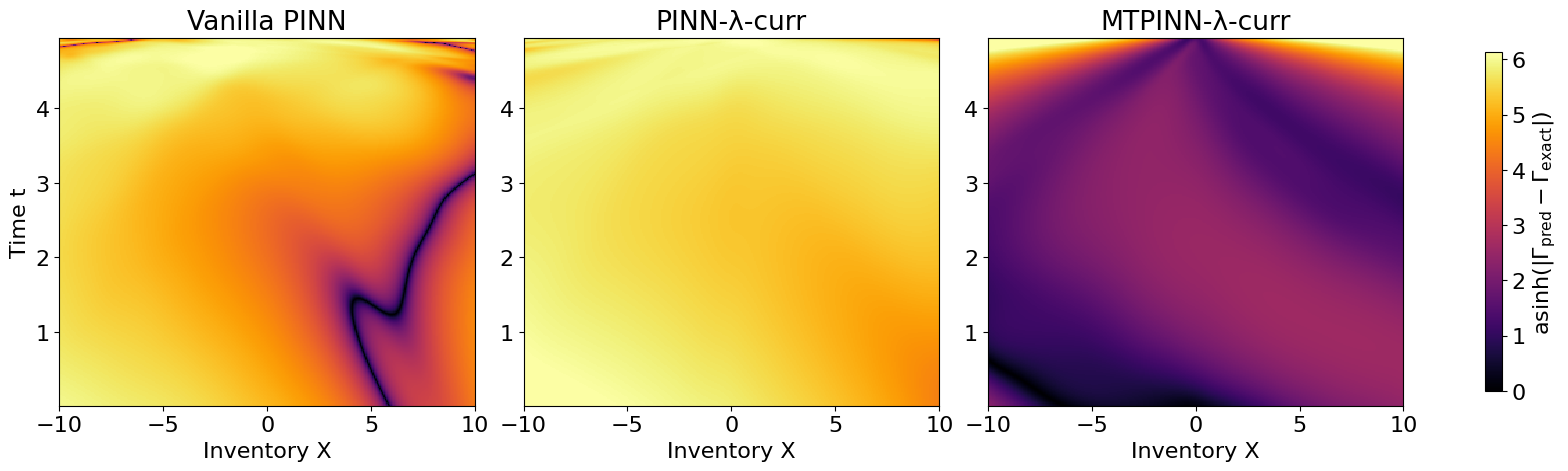}
    \caption{$\lambda=0.10$}
  \end{subfigure}

  \caption{Heatmaps of domain-wide error over \((t,X)\) for each risk aversion \(\lambda\) with fixed $S = 55$.
  Each panel compares methods as in the per-\(\lambda\) heatmap export.}
  \label{fig:heatmaps-by-lambda}
\end{figure}

\begin{figure}
  \centering
  \begin{subfigure}[t]{0.75\linewidth}
    \centering\includegraphics[width=7cm]{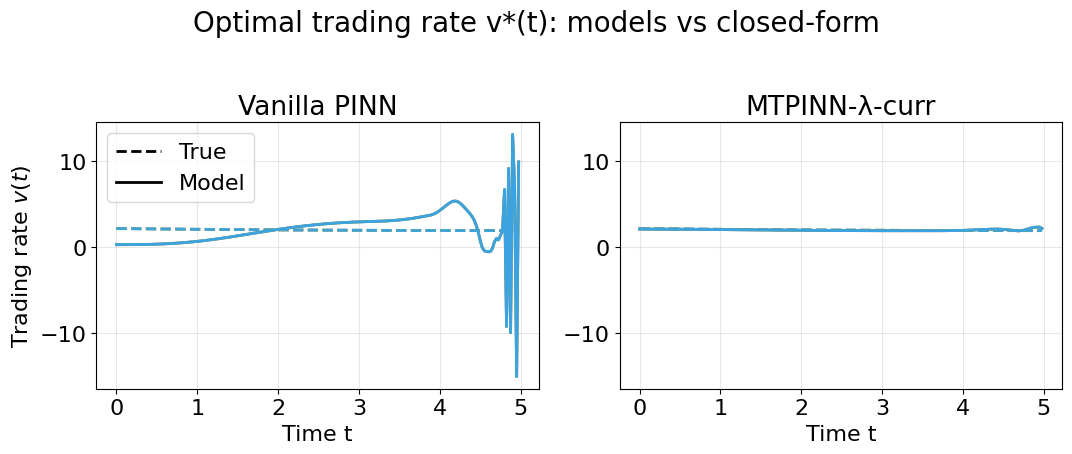}
    \caption{$\lambda=0$}
  \end{subfigure}\hfill
  \begin{subfigure}[t]{0.75\linewidth}
    \centering\includegraphics[width=\linewidth]{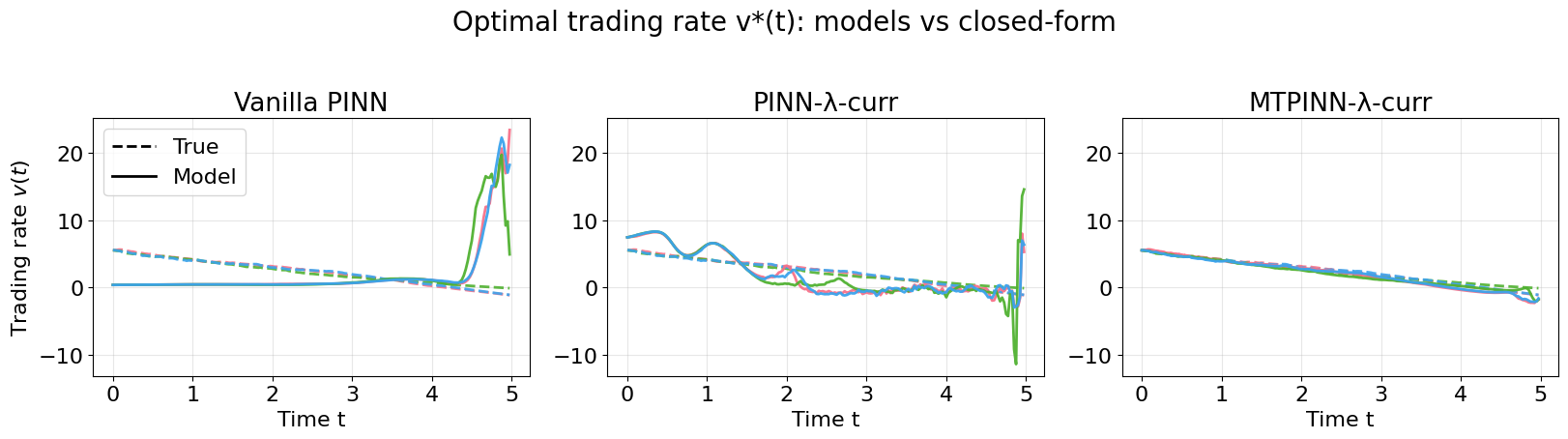}
    \caption{$\lambda=0.05$}
  \end{subfigure}\hfill
  \begin{subfigure}[t]{0.75\linewidth}
    \centering\includegraphics[width=\linewidth]{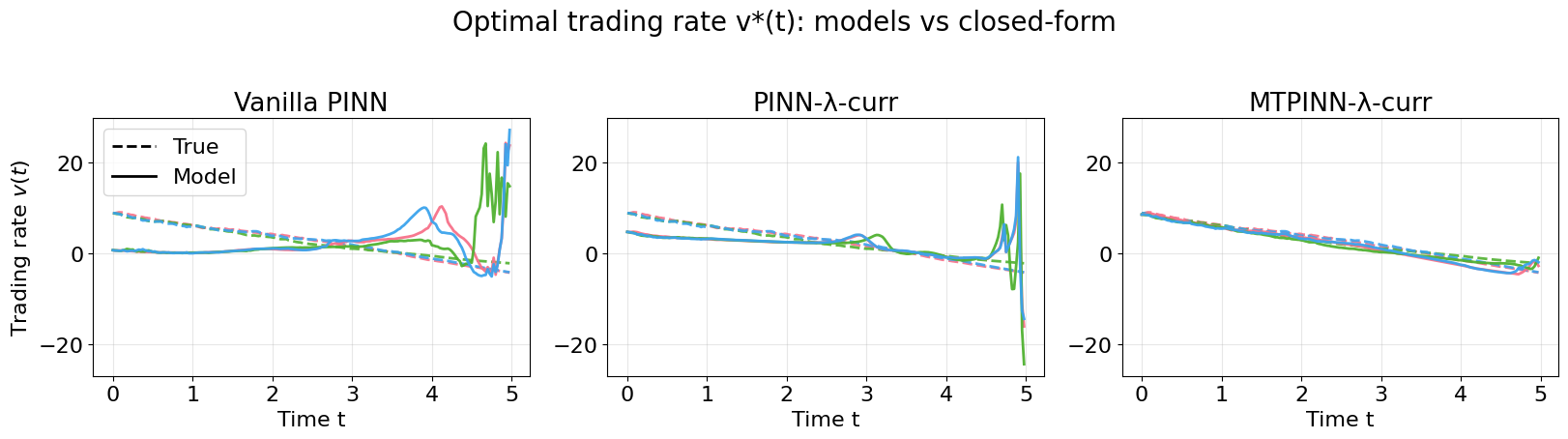}
    \caption{$\lambda=0.10$}
  \end{subfigure}
  \caption{Optimal trading rate $v^*(t)$: model vs. closed form for each $\lambda$.}
  \label{fig:trading-rate-by-lambda}
\end{figure}

\begin{figure}
  \centering
  \begin{subfigure}[t]{0.55\linewidth}
    \centering\includegraphics[width=7cm]{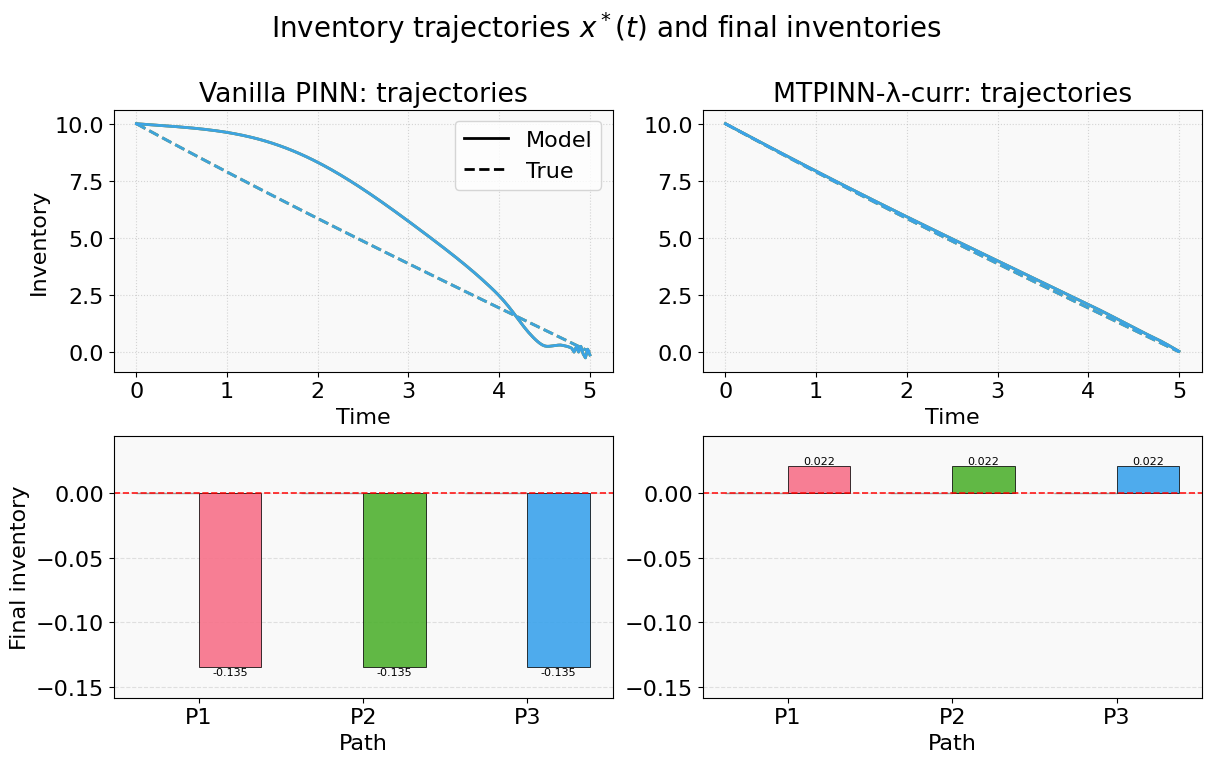}
    \caption{$\lambda=0$}
  \end{subfigure}\hfill
  \begin{subfigure}[t]{0.55\linewidth}
    \centering\includegraphics[width=\linewidth]{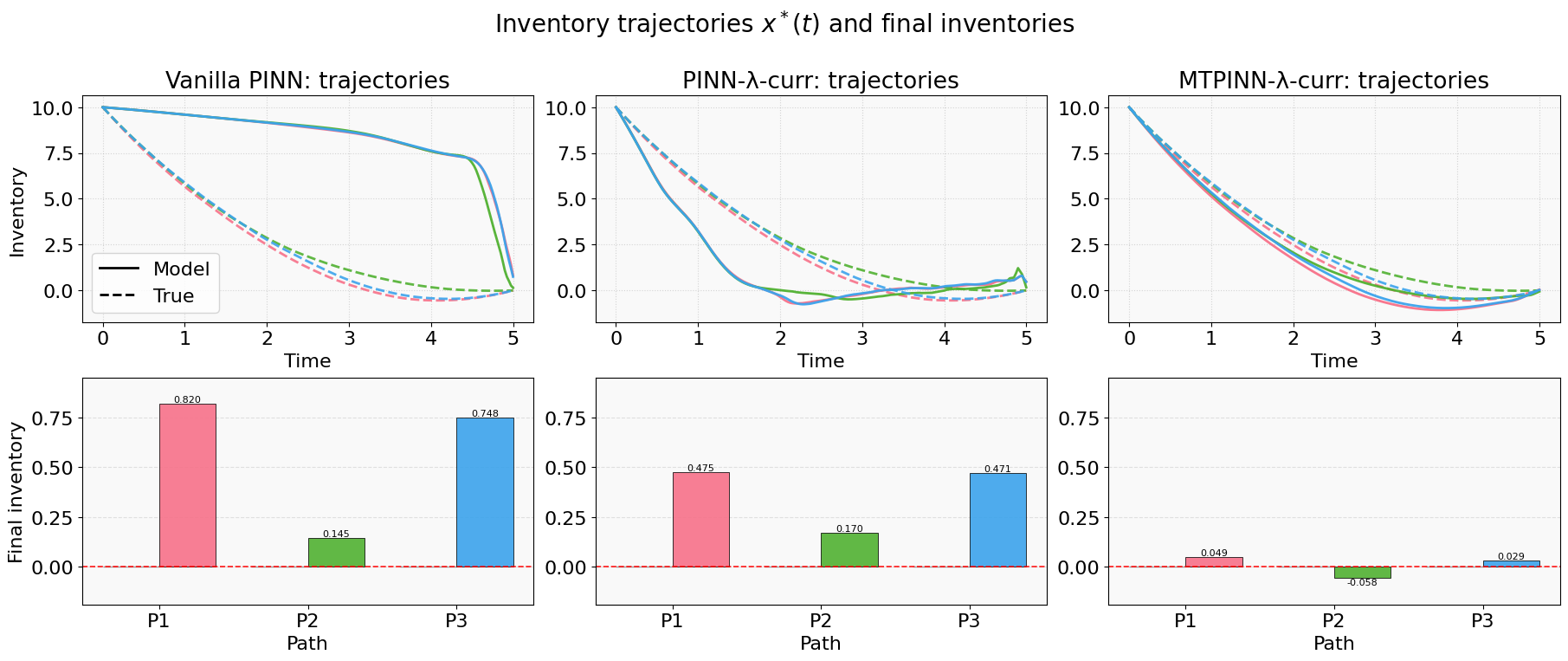}
    \caption{$\lambda=0.05$}
  \end{subfigure}\hfill
  \begin{subfigure}[t]{0.55\linewidth}
    \centering\includegraphics[width=\linewidth]{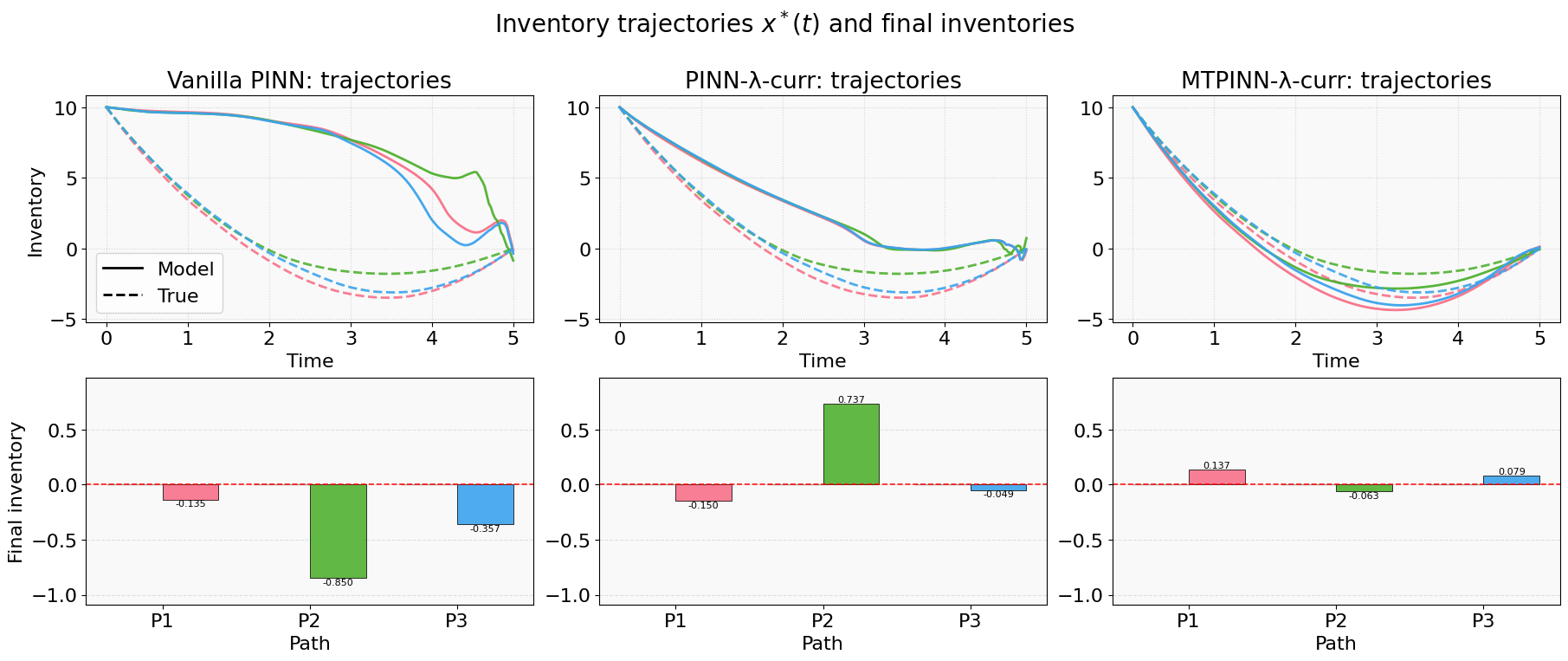}
    \caption{$\lambda=0.10$}
  \end{subfigure}
  \caption{Inventory trajectories $x^*(t)$ and final inventories across paths for each $\lambda$.}
  \label{fig:traj-by-lambda}
\end{figure}

\begin{figure}
  \centering
  \begin{subfigure}[t]{0.5\linewidth}
    \centering\includegraphics[width=\linewidth]{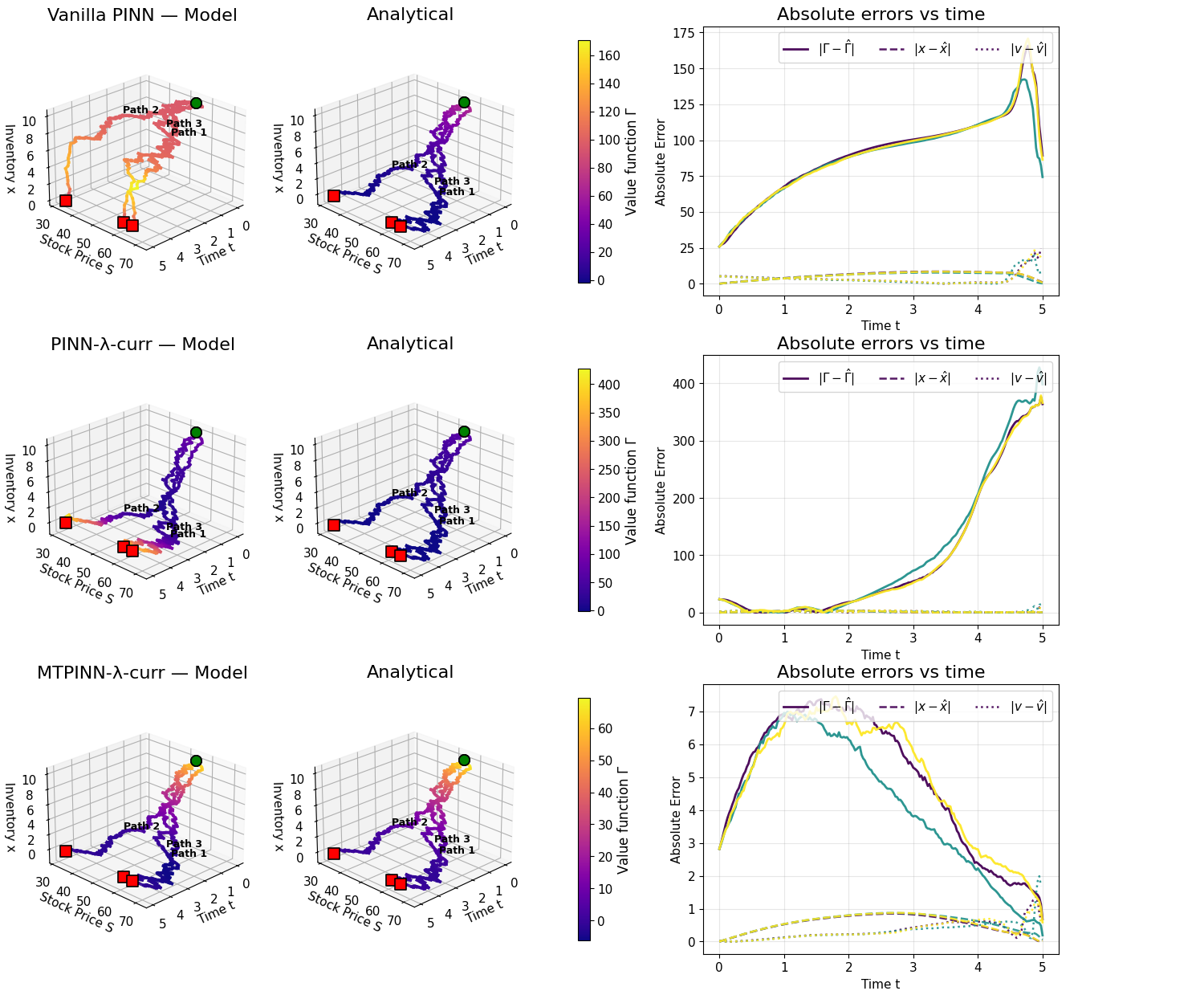}
    \caption{$\lambda=0.05$}
  \end{subfigure}\hfill
  \begin{subfigure}[t]{0.5\linewidth}
    \centering\includegraphics[width=\linewidth]{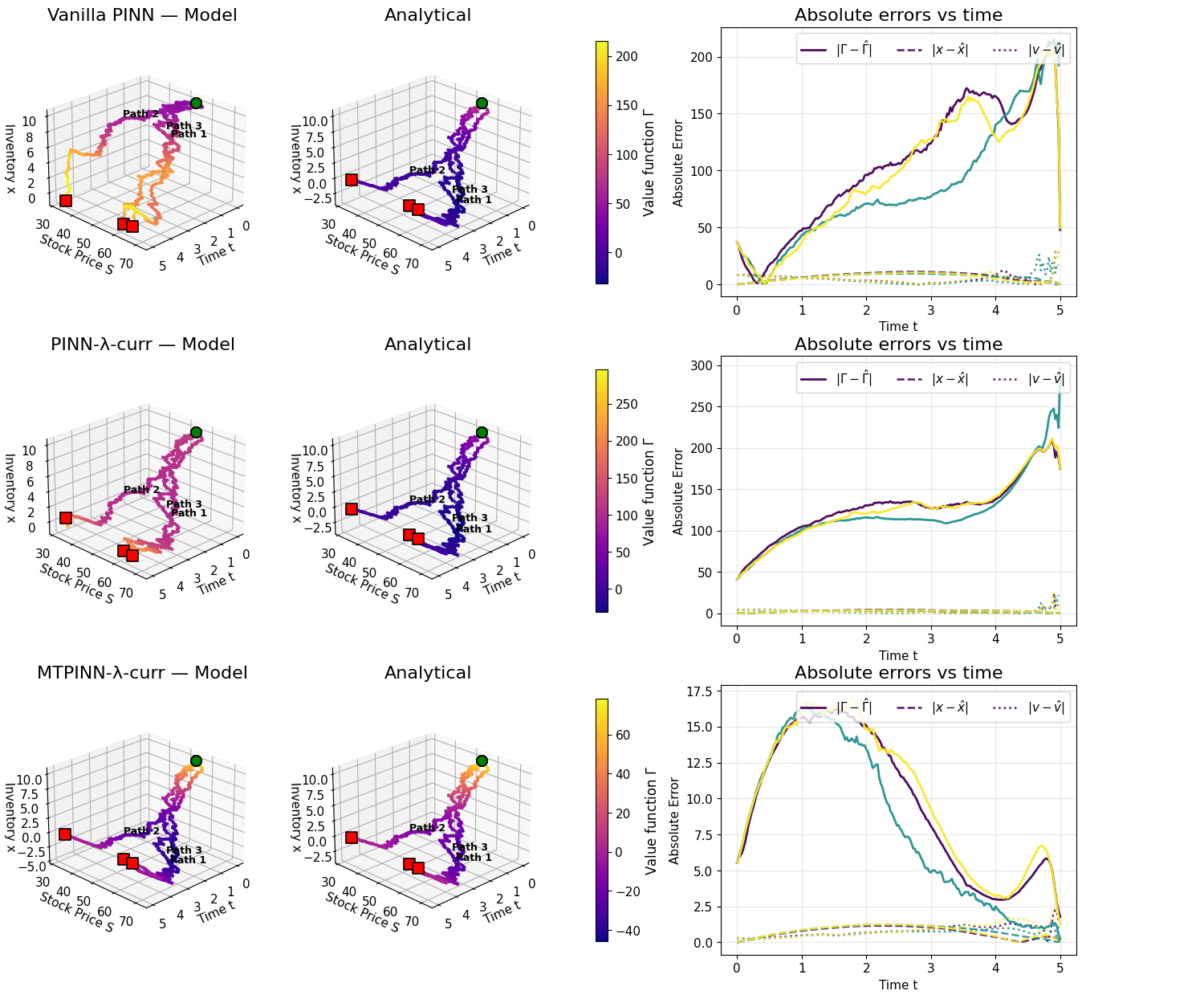}
    \caption{$\lambda=0.10$}
  \end{subfigure}
  \caption{Pathwise rollouts: model vs.\ analytic trajectories and absolute error over time for each $\lambda > 0$.}
  \label{fig:paths-by-lambda}
\end{figure}

\begin{figure*}[t]
  \centering
  % Row: λ = 0
  \begin{subfigure}[t]{0.32\textwidth}
    \centering\includegraphics[width=\linewidth]{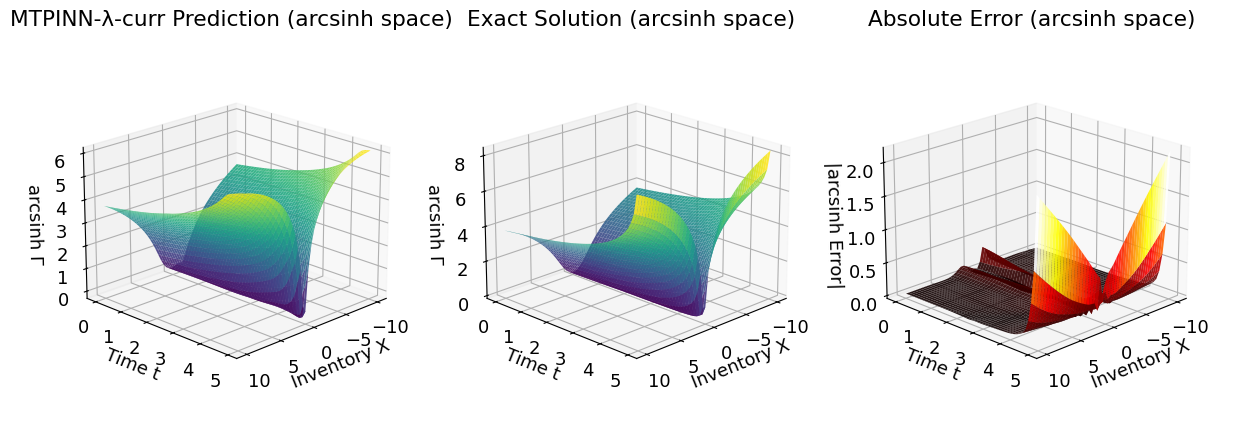}
    \caption{MT-PINN-$\lambda$-curr, $\lambda=0$}
  \end{subfigure}\hfill
  \begin{subfigure}[t]{0.32\textwidth}
    \centering\includegraphics[width=\linewidth]{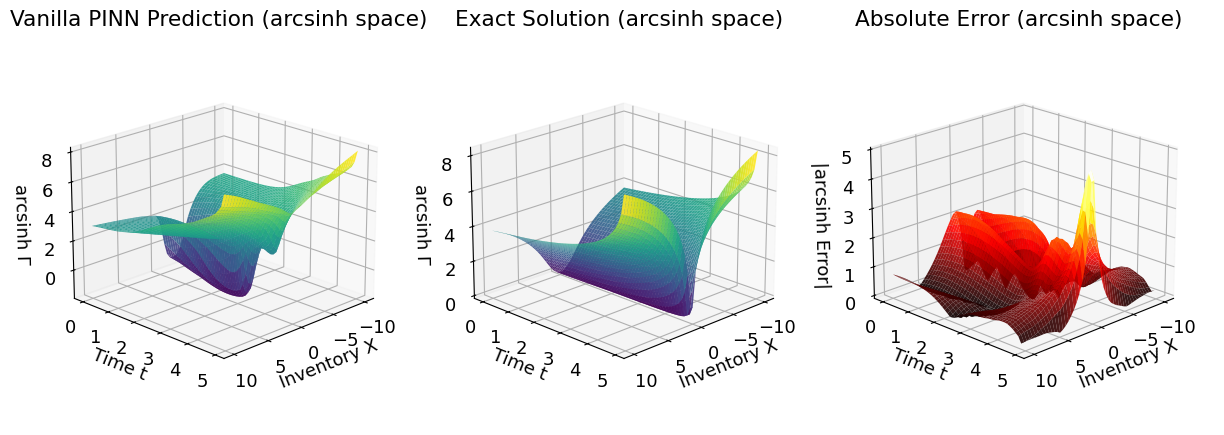}
    \caption{Vanilla PINN, $\lambda=0$}
  \end{subfigure}

  % Row: λ = 0.05
  \par\medskip
  \begin{subfigure}[t]{0.32\textwidth}
    \centering\includegraphics[width=\linewidth]{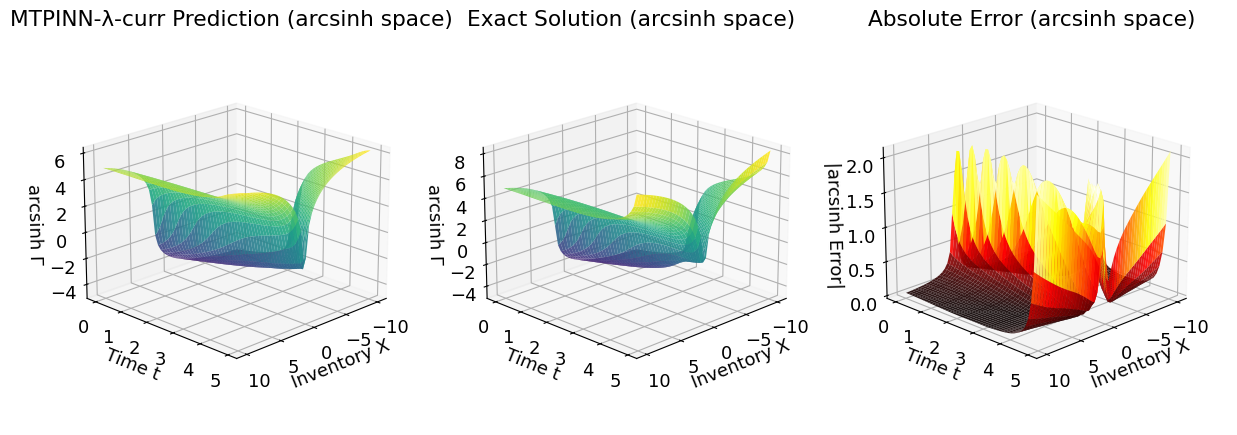}
    \caption{MT-PINN-$\lambda$-curr, $\lambda=0.05$}
  \end{subfigure}\hfill
  \begin{subfigure}[t]{0.32\textwidth}
    \centering\includegraphics[width=\linewidth]{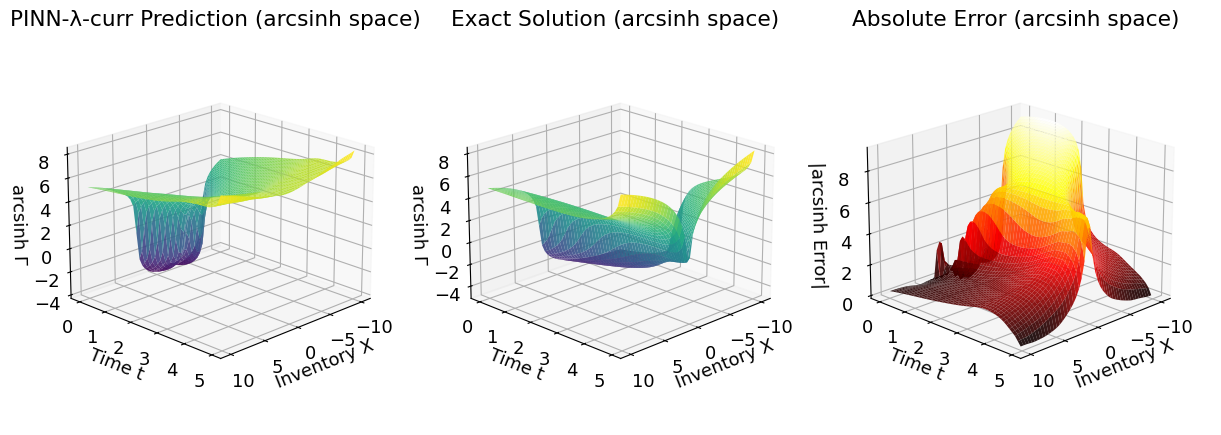}
    \caption{PINN-$\lambda$-curr, $\lambda=0.05$}
  \end{subfigure}\hfill
  \begin{subfigure}[t]{0.32\textwidth}
    \centering\includegraphics[width=\linewidth]{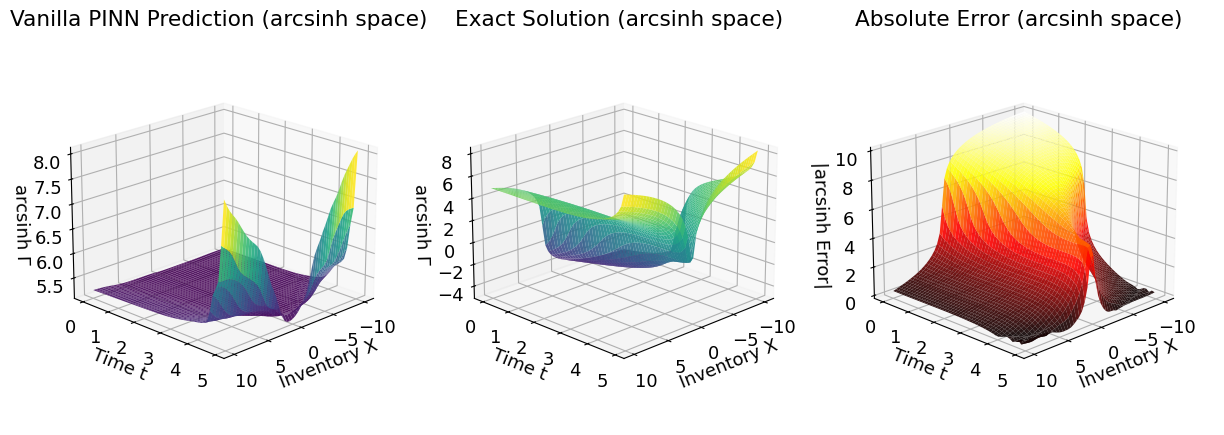}
    \caption{Vanilla PINN, $\lambda=0.05$}
  \end{subfigure}

  % Row: λ = 0.10
  \par\medskip
  \begin{subfigure}[t]{0.32\textwidth}
    \centering\includegraphics[width=\linewidth]{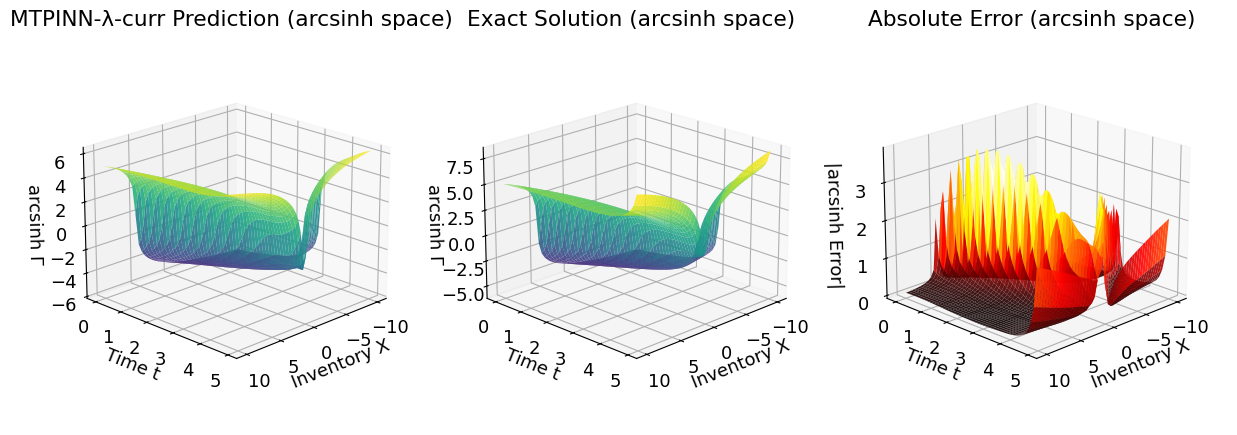}
    \caption{MT-PINN-$\lambda$-curr, $\lambda=0.10$}
  \end{subfigure}\hfill
  \begin{subfigure}[t]{0.32\textwidth}
    \centering\includegraphics[width=\linewidth]{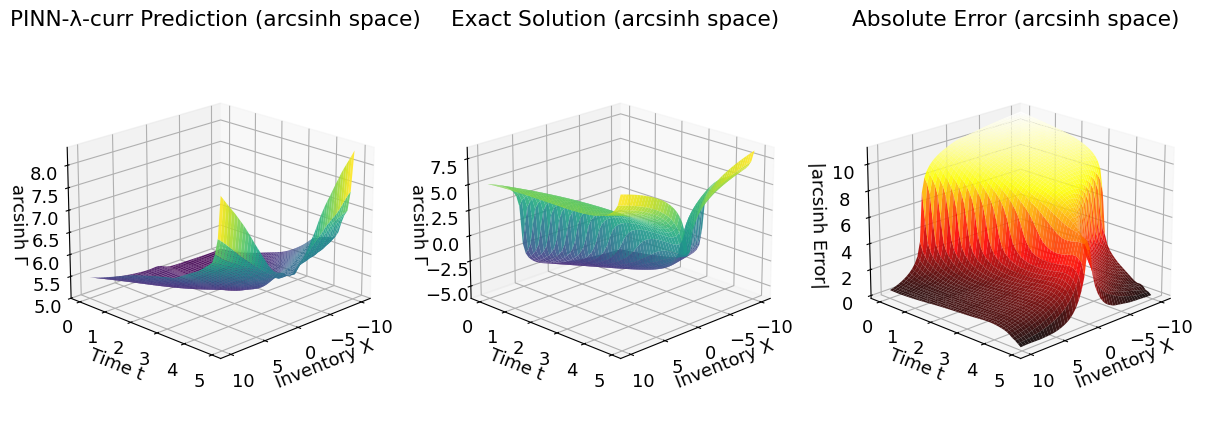}
    \caption{PINN-$\lambda$-curr, $\lambda=0.10$}
  \end{subfigure}\hfill
  \begin{subfigure}[t]{0.32\textwidth}
    \centering\includegraphics[width=\linewidth]{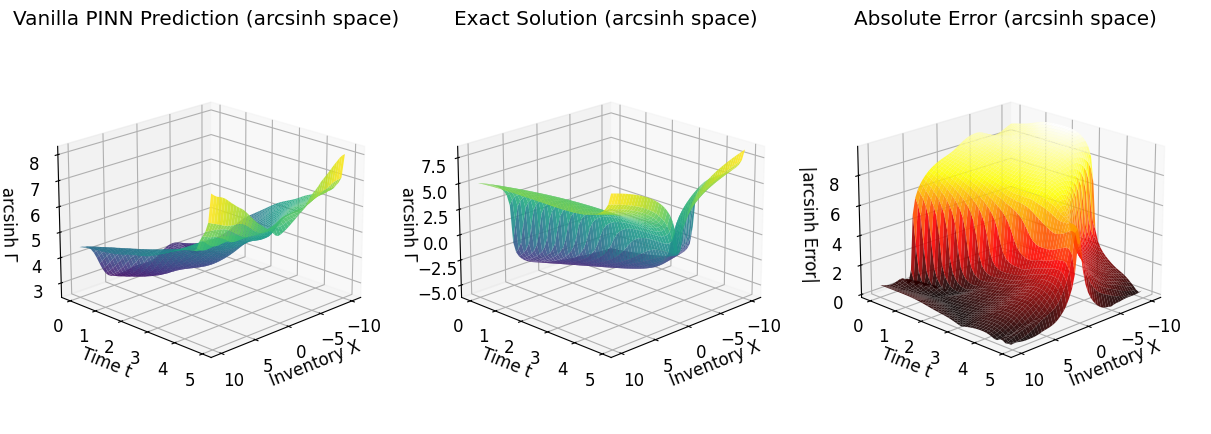}
    \caption{Vanilla PINN, $\lambda=0.10$}
  \end{subfigure}
  \caption{Arcsinh-space value surfaces (each panel is organized as: Prediction / Exact / Absolute Error). Rows vary $\lambda$, columns vary the method.}
  \label{fig:asinh-surfaces-3x3}
\end{figure*}

\begin{table}
  \caption{Absolute terminal-inventory enforcement on the synthetic benchmark for 200 simulated price paths for varying $\lambda$. It is reported $p_\varepsilon=\Pr(|X_T|\le\varepsilon)$ with $\varepsilon=0.05$.}
  \label{tab:terminal-enforcement-all-lam}
  \centering
  \setlength{\tabcolsep}{3pt}     % tighter columns
  \renewcommand{\arraystretch}{0.9}
  \scriptsize
  \begin{tabular}{l ccc ccc ccc}
    \toprule
    & \multicolumn{3}{c}{$\lambda=0$} & \multicolumn{3}{c}{$\lambda=0.05$} & \multicolumn{3}{c}{$\lambda=0.10$} \\
    \cmidrule(lr){2-4}\cmidrule(lr){5-7}\cmidrule(lr){8-10}
    Method & Mean$\pm$Std & 95th pct & $p_\varepsilon$
           & Mean$\pm$Std & 95th pct & $p_\varepsilon$
           & Mean$\pm$Std & 95th pct & $p_\varepsilon$ \\
    \midrule
    Vanilla PINN            & 0.135$\pm$0.000 & 0.135 & 0.000 & 0.570$\pm$0.196 & 0.826 & 0.000 & 0.777$\pm$0.444 & 1.407 & 0.055 \\
    PINN-$\lambda$-curr     & - & - & - & 0.420$\pm$0.085 & 0.481 & 0.000 & 0.164$\pm$0.161 & 0.527 & 0.205 \\
    MT-PINN-$\lambda$-curr  & \textbf{0.022$\pm$0.000} & \textbf{0.022} & \textbf{1.000}
                             & \textbf{0.031$\pm$0.030} & \textbf{0.072} & \textbf{0.810}
                             & \textbf{0.073$\pm$0.092} & \textbf{0.241} & \textbf{0.600} \\
    \bottomrule
  \end{tabular}
\end{table}

\begin{table*}[htbp]
\centering
 \caption{Error statistics for Vanilla PINN, PINN-$\lambda$-curr, and MT-PINN-$\lambda$-curr across $\lambda \in \{0,0.05,0.10\}$ (with PINN-$\lambda$-curr omitted at $\lambda=0$) and fixed stock price $S = 55$. Mean absolute error, max absolute error, mean relative error, and root-mean squared error are reported in both original space and arcsinh-transformed space.}
\label{tab:surface-error-stats}
\resizebox{\textwidth}{!}{%
\begin{tabular}{llcccccccc}
\toprule
\multirow{2}{*}{$\lambda$} & \multirow{2}{*}{Method} & \multicolumn{4}{c}{Original Space} & \multicolumn{4}{c}{Arcsinh Space} \\
\cmidrule(lr){3-6} \cmidrule(lr){7-10}
 & & MAE & MaxAE & MRE & RMSE & MAE & MaxAE & MRE & RMSE \\
\midrule
\multirow{3}{*}{0} 
 & Vanilla PINN   & 22.75 & \textbf{363.18} & 7.43  & \textbf{45.45} & 0.80 & 4.96 & 3.66 & 1.07 \\
 & PINN-$\lambda$-curr & - & - & - & - & - & - & - & - \\
 & MT-PINN-$\lambda$-curr    & \textbf{16.44} & 1773.53 & \textbf{0.24}  & 116.00 & \textbf{0.12} & \textbf{2.18} & \textbf{0.18} & \textbf{0.25} \\
\midrule
\multirow{3}{*}{0.05} 
 & Vanilla    & 78.98 & 573.98 & 33.69 & \textbf{88.12} & 3.97 & 10.03 & 2.80 & 5.29 \\
 & PINN-$\lambda$-curr & 145.77 & \textbf{413.17} & 62.52 & 192.12 & 3.49 & 9.30 & 2.66 & 4.51 \\
 & MT-PINN-$\lambda$-curr    & \textbf{17.73} & 1756.74 & \textbf{0.77}  & 115.68 & \textbf{0.28} & \textbf{2.10} & \textbf{0.51} & \textbf{0.50} \\
\midrule
\multirow{3}{*}{0.10} 
 & Vanilla PINN   & 81.22 & 475.45 & 33.92 & \textbf{99.78} & 4.33 & 9.71 & 2.27 & 5.61 \\
 & PINN-$\lambda$-curr & 140.12 & \textbf{270.59} & 44.04 & 147.14 & 5.23 & 11.02 & 2.60 & 6.57 \\
 & MT-PINN-$\lambda$-curr    & \textbf{19.65} & 1742.84 & \textbf{1.09}  & 115.39 & \textbf{0.37} & \textbf{3.84} & \textbf{0.56} & \textbf{0.73} \\
\bottomrule
\end{tabular}%
}
\end{table*}

For $\lambda=0$ only vanilla PINN and MT-PINN-$\lambda$-curriculum are shown as PINN-$\lambda$-curriculum collapses to vanilla PINN and is omitted.

Figure \ref{fig:final-inventory-dists} shows the histogram of the absolute terminal inventory $|X_T|$ across 200 stock price simulations with an $\epsilon$ tolerance market and mean±std inset of the terminal inventory. It is evident that the MT-PINN-$\lambda$-curriculum concentrates heavily near zero and attains the highest pass-rate $p_\epsilon = Pr(|X_T| \le \epsilon)$ for every $\lambda$ (see Table \ref{tab:terminal-enforcement-all-lam}). Baselines show larger violations for zero terminal inventory, with the PINN-$\lambda$-curriculum exhibiting improvements over the vanilla PINN for larger $\lambda$.

Figure \ref{fig:trading-rate-by-lambda} illustrates the implied trading rate $v^*(t)=\frac{1}{2}\partial_X \Gamma_\theta$ versus the analytical solution derived by Gatheral and Schied for all three paths. MT-PINN-$\lambda$-curriculum follows closer to the closed-form across the horizon for all $\lambda$, while the baselines show higher variance. 

Figure \ref{fig:traj-by-lambda} reveals the inventory trajectories $x^*(t)$ under the learned trading rate for all three paths. Although MT-PINN-$\lambda$-curriculum's accuracy in matching the closed-form solution reduces as $\lambda$ increases, it consistently produces near zero-inventory at terminal and nevertheless outperforms the baseline PINNs. The PINN-$\lambda$-curriculum improves on the vanilla PINN for higher $\lambda$, somewhat grasping the concept of reduced risk exposure, unlike the vanilla PINN which increases its risk exposure.

Figure \ref{fig:paths-by-lambda} displays the models versus analytical inventory trajectories for three paths, showing the value function along the path and the absolute error over time for the value function, inventory trajectory, and trading rate. Note that the plots only show for $\lambda > 0$, because, as previously stated, for $\lambda=0$, the optimal control and trajectories are invariant to the stock price $S$. MT-PINN-$\lambda$-curriculum shows a reduction in nearer terminal errors, reduced path-variance, and orders of magnitude smaller error than the baselines. 

Figure \ref{fig:heatmaps-by-lambda} exhibits the domain-wide inverse hyperbolic sine function absolute value function error $(\text{asinh}(|\Gamma_\text{pred} - \Gamma_\text{exact}|))$ over $(t, X)$ for fixed mid-range stock price $S=55$. MT-PINN-$\lambda$-curriculum has reduced error across the domain with higher error near terminal, as can be seen in Table \ref{tab:surface-error-stats}. Table \ref{tab:surface-error-stats} shows that across the arcsinh space, MT-PINN-$\lambda$-curriculum performs the best across all metrics. In the original space, MT-PINN-$\lambda$-curriculum performs best in mean absolute error and mean-relative error, while PINN-$\lambda$-curriculum consistently performed best in root-mean-squared error and, for the most part, maximum absolute error. The rationale for why MT-PINN-$\lambda$-curriculum performs better in the arcsinh space than it does in the original space is because the errors in the outliers, mostly found near maturity, aren't penalized as much in the arcsinh space. This demonstrates that the baseline PINNs capture the singularity in the value function with less error, however, overall, MT-PINN-$\lambda$-curriculum better captures the value function across the entire domain. 

Figure \ref{fig:asinh-surfaces-3x3} provides a 3D view of Figure \ref{fig:heatmaps-by-lambda}. It is evident that, domain-wide, the baseline PINNs have less error near terminal than the MT-PINN-$\lambda$-curriculum due to their terminal condition loss; however, as has been outlined, this doesn't result in accurate control policy.

\subsection{SPY Real-Market Backtest Results}\label{Appendix F.2}
\FloatBarrier

\begin{figure}[t]
    \centering
    \includegraphics[width=0.75\linewidth]{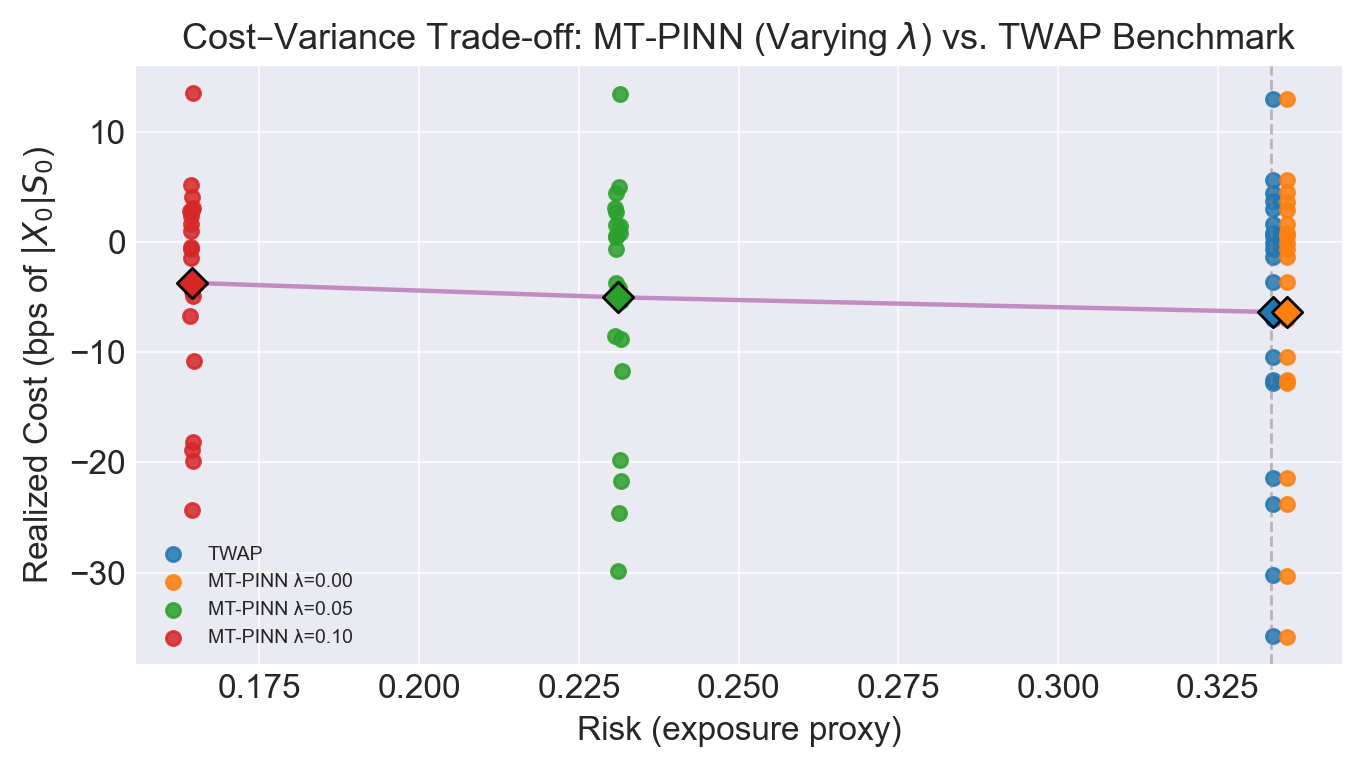}
    \caption{Cost-variance/exposure trade-off of MT-PINN for varying $\lambda \in [0,0.05,0.1]$ and TWAP.}
    \label{fig:cost-variance}
\end{figure}

\begin{figure}[t]
    \centering
    \includegraphics[width=1.0\linewidth]{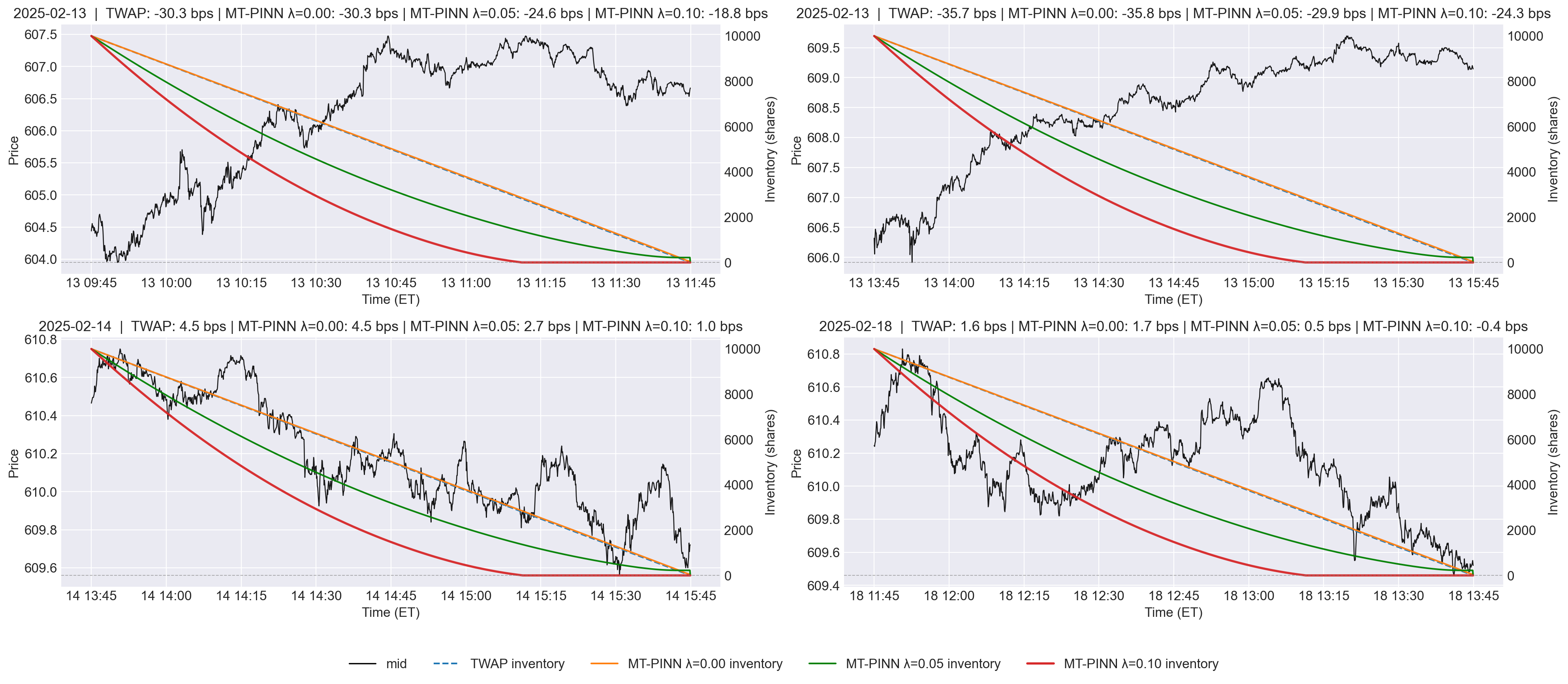}
    \caption{Execution trajectories and costs for MT-PINN with varying $\lambda \in [0,0.05,0.1]$ versus TWAP over four sampled trading windows of the SPY $\left ([W_{10}, W_{12}, W_{15}, W_{17}] \right )$. The top two subplots show rising SPY windows, while the bottom two subplots show falling SPY windows.}
    \label{fig:window_overlay}
\end{figure}

\begin{figure}
  \centering
  \begin{subfigure}[t]{0.32\linewidth}
    \centering
    \includegraphics[width=\linewidth]{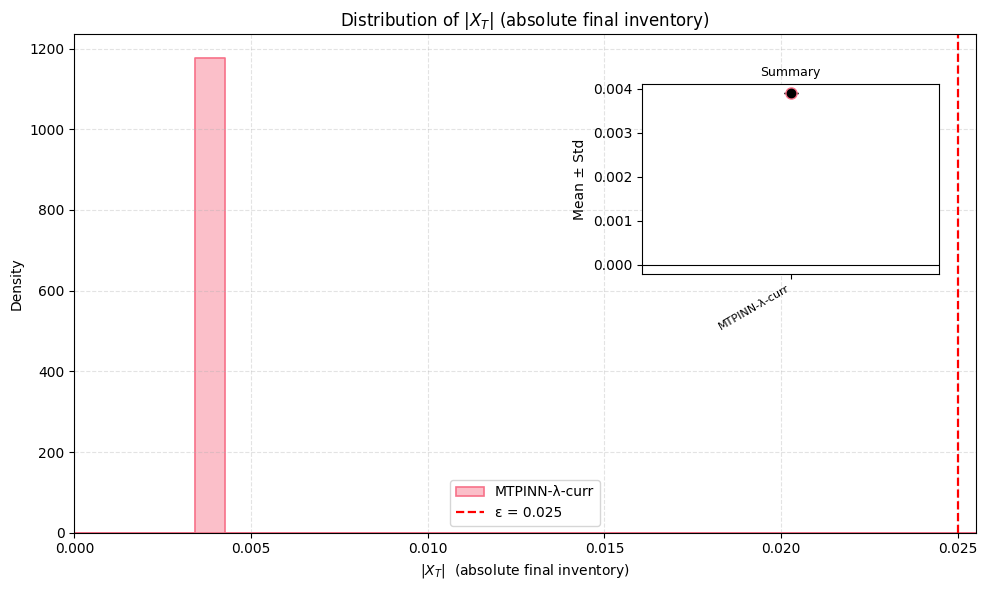}%
    \caption{$\lambda=0$}
  \end{subfigure}\hfill
  \begin{subfigure}[t]{0.32\linewidth}
    \centering
    \includegraphics[width=\linewidth]{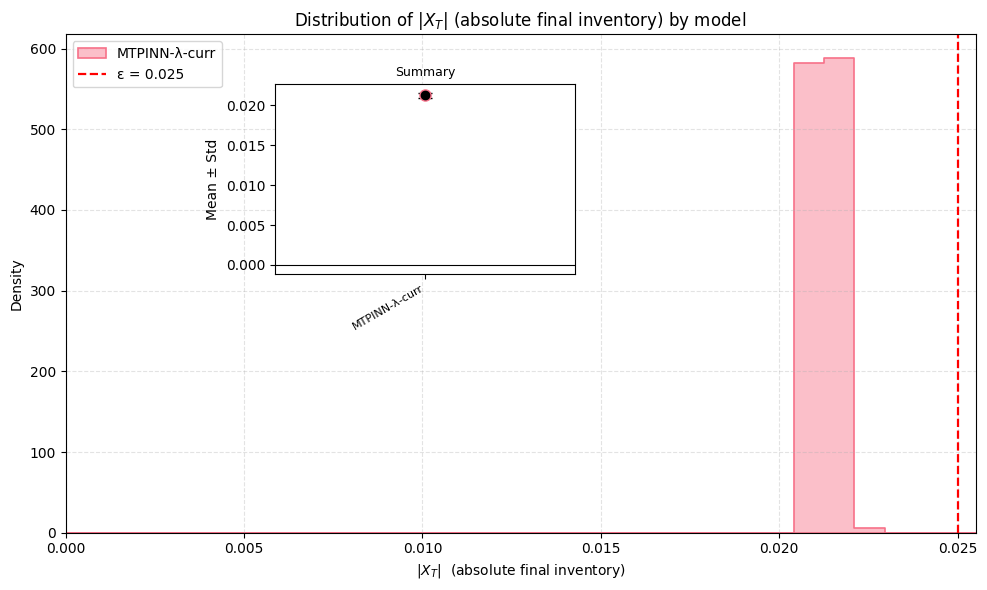}%
    \caption{$\lambda=0.05$}
  \end{subfigure}\hfill
  \begin{subfigure}[t]{0.32\linewidth}
    \centering
    \includegraphics[width=\linewidth]{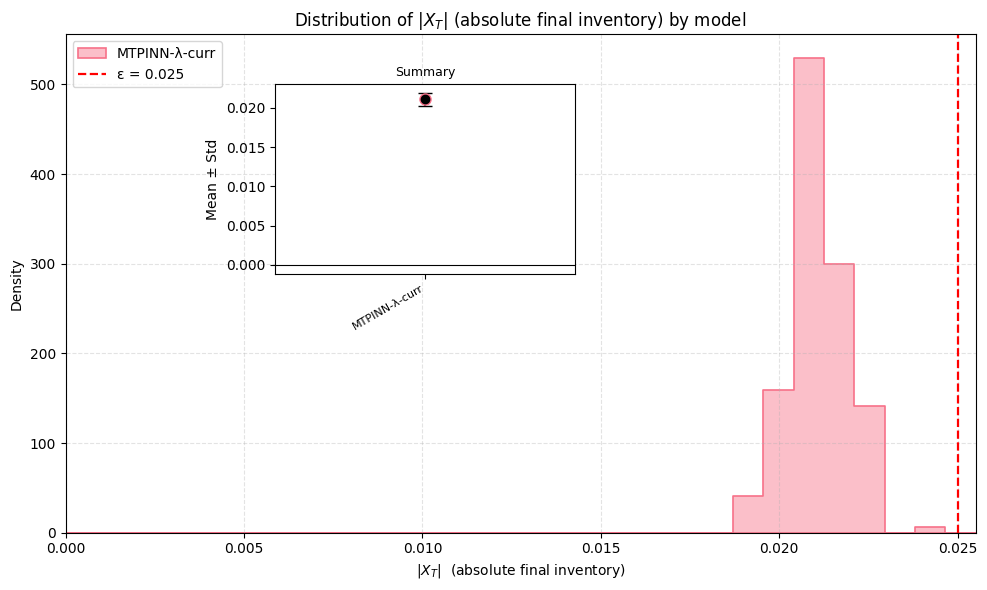}%
    \caption{$\lambda=0.10$}
  \end{subfigure}

  \caption{Distribution of \(|X_T|\) (absolute final inventory) for each risk aversion \(\lambda\).
  Each panel shows the histogram across rollouts for the MT-PINN model with the tolerance line at \(\epsilon=0.025\) and the inset reporting mean\(\,\pm\)std.
  }
  \label{fig:final-inventory-dists-market}
\end{figure}

\begin{figure}
  \centering
  \begin{subfigure}[t]{0.32\linewidth}
    \centering
    \includegraphics[width=\linewidth]{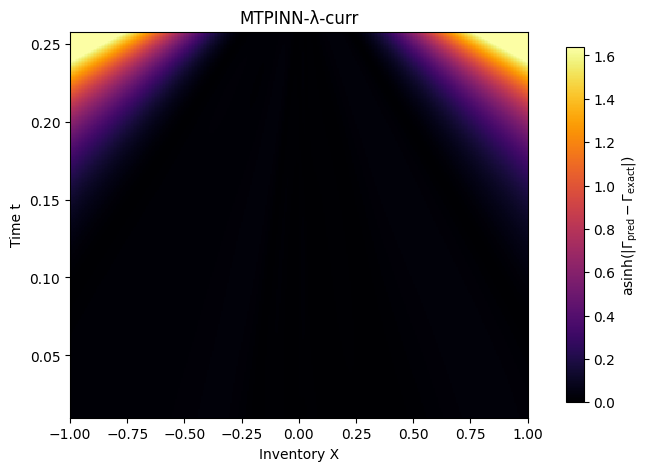}%
    \caption{$\lambda=0$}
  \end{subfigure}\hfill
  \begin{subfigure}[t]{0.32\linewidth}
    \centering
    \includegraphics[width=\linewidth]{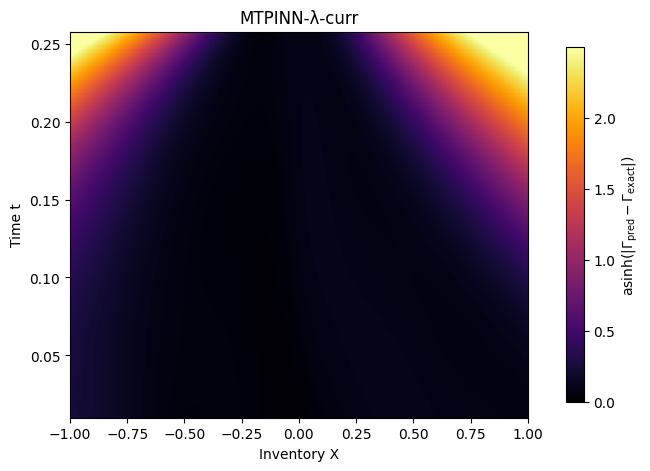}%
    \caption{$\lambda=0.05$}
  \end{subfigure}\hfill
  \begin{subfigure}[t]{0.32\linewidth}
    \centering
    \includegraphics[width=\linewidth]{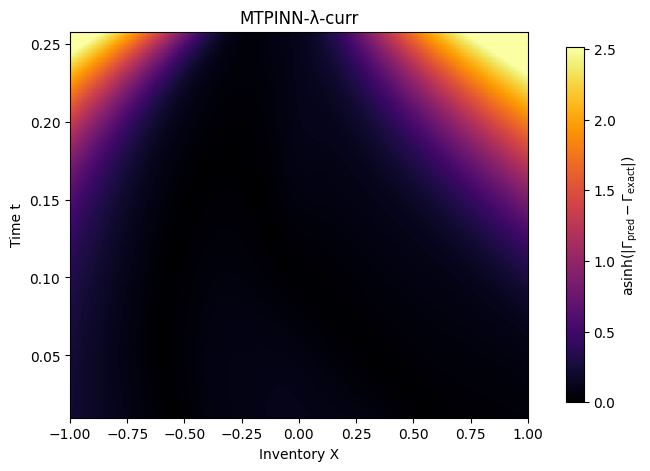}%
    \caption{$\lambda=0.10$}
  \end{subfigure}

  \caption{Heatmaps of domain-wide error over $(t, X)$ for each risk aversion $\lambda$ with fixed $S= 605$.
  }
  \label{fig:heatmap-market}
\end{figure}

\begin{figure}
  \centering
  \begin{subfigure}[t]{0.32\linewidth}
    \centering
    \includegraphics[width=\linewidth]{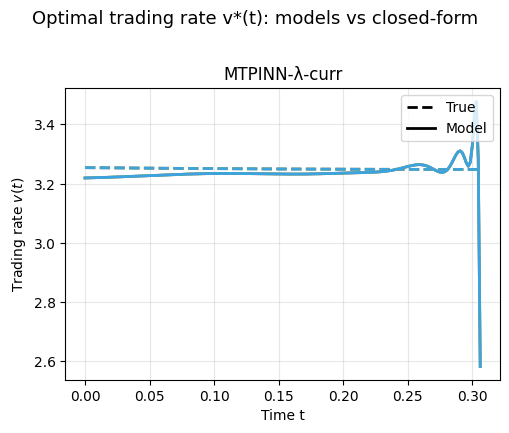}%
    \caption{$\lambda=0$}
  \end{subfigure}\hfill
  \begin{subfigure}[t]{0.32\linewidth}
    \centering
    \includegraphics[width=\linewidth]{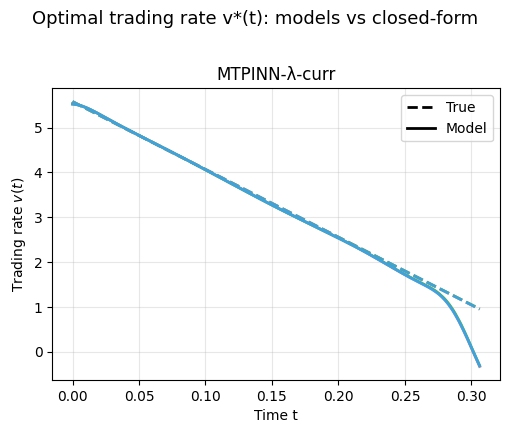}%
    \caption{$\lambda=0.05$}
  \end{subfigure}\hfill
  \begin{subfigure}[t]{0.32\linewidth}
    \centering
    \includegraphics[width=\linewidth]{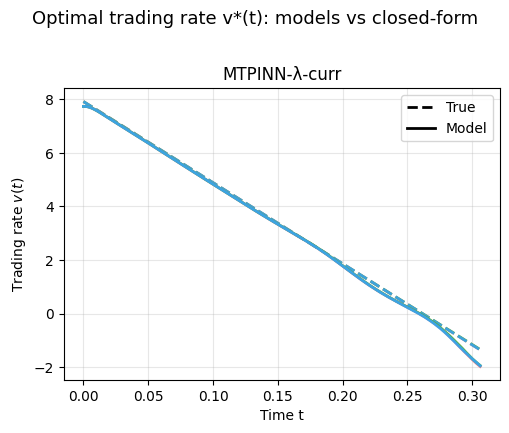}%
    \caption{$\lambda=0.10$}
  \end{subfigure}

  \caption{Optimal trading rate $v^*(t)$: model vs. closed form for each $\lambda$.
  }
  \label{fig:trading-rate-market}
\end{figure}

\begin{figure}
  \centering
  \begin{subfigure}[t]{0.32\linewidth}
    \centering
    \includegraphics[width=\linewidth]{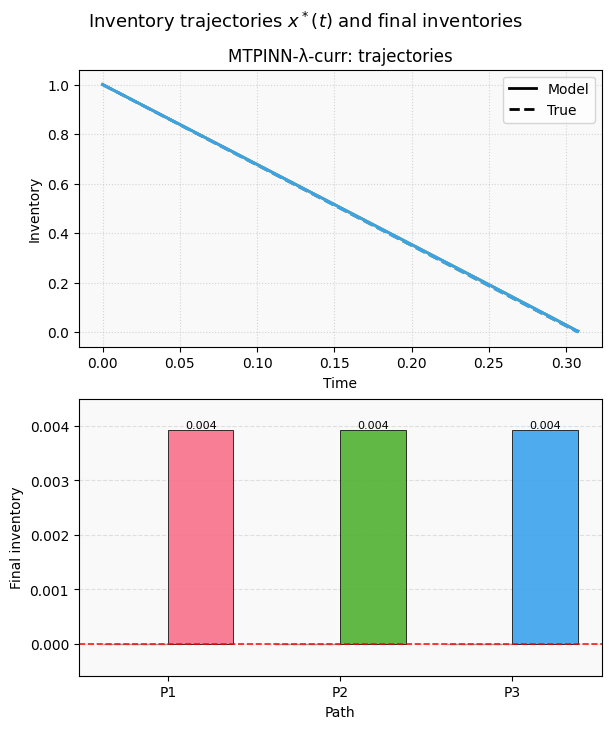}%
    \caption{$\lambda=0$}
  \end{subfigure}\hfill
  \begin{subfigure}[t]{0.32\linewidth}
    \centering
    \includegraphics[width=\linewidth]{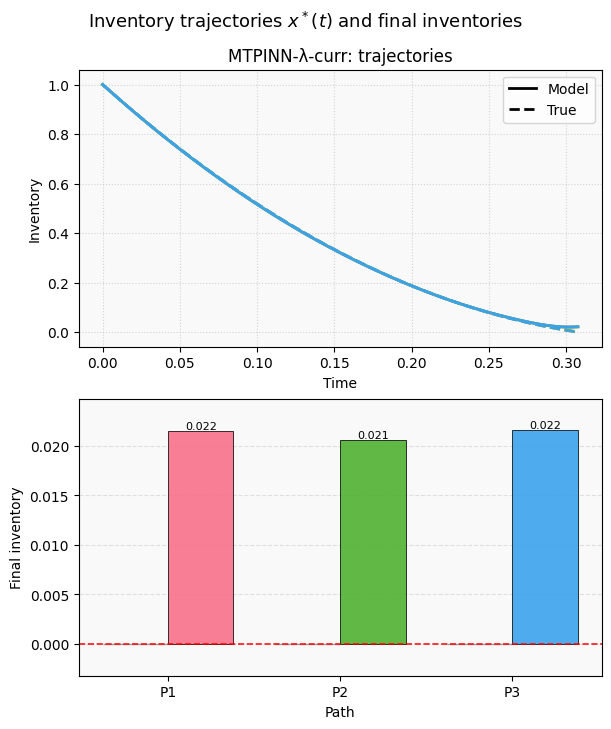}%
    \caption{$\lambda=0.05$}
  \end{subfigure}\hfill
  \begin{subfigure}[t]{0.32\linewidth}
    \centering
    \includegraphics[width=\linewidth]{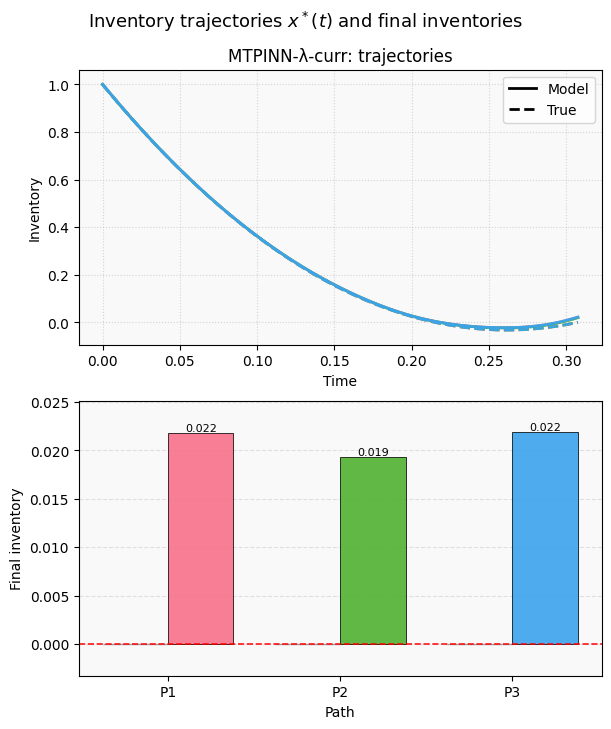}%
    \caption{$\lambda=0.10$}
  \end{subfigure}

  \caption{Inventory trajectories $x^*(t)$ and final inventories across paths for each $\lambda$.}
  \label{fig:inventory-traj-market}
\end{figure}

\begin{figure}
  \centering
  \begin{subfigure}[t]{0.45\linewidth}
    \centering
    \includegraphics[width=\linewidth]{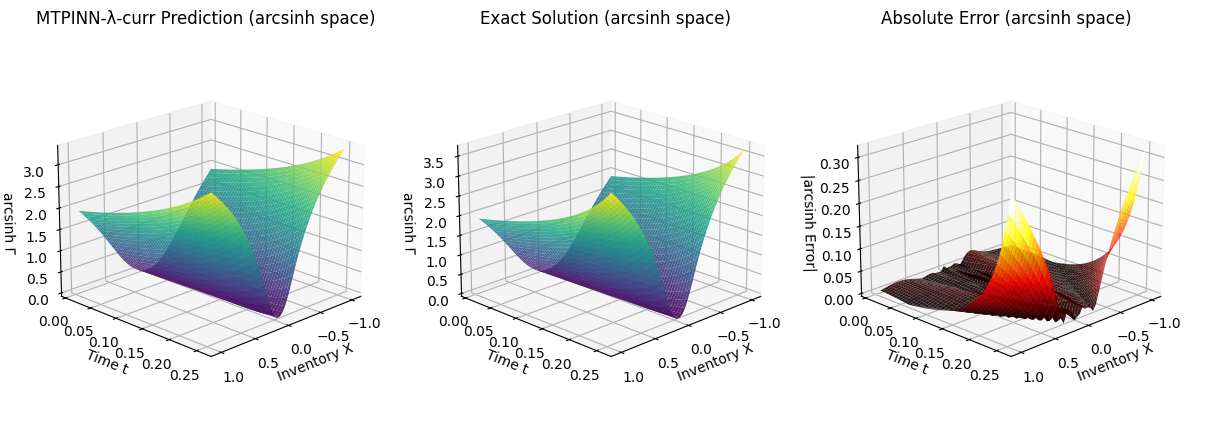}%
    \caption{$\lambda=0$}
  \end{subfigure}\hfill
  \begin{subfigure}[t]{0.45\linewidth}
    \centering
    \includegraphics[width=\linewidth]{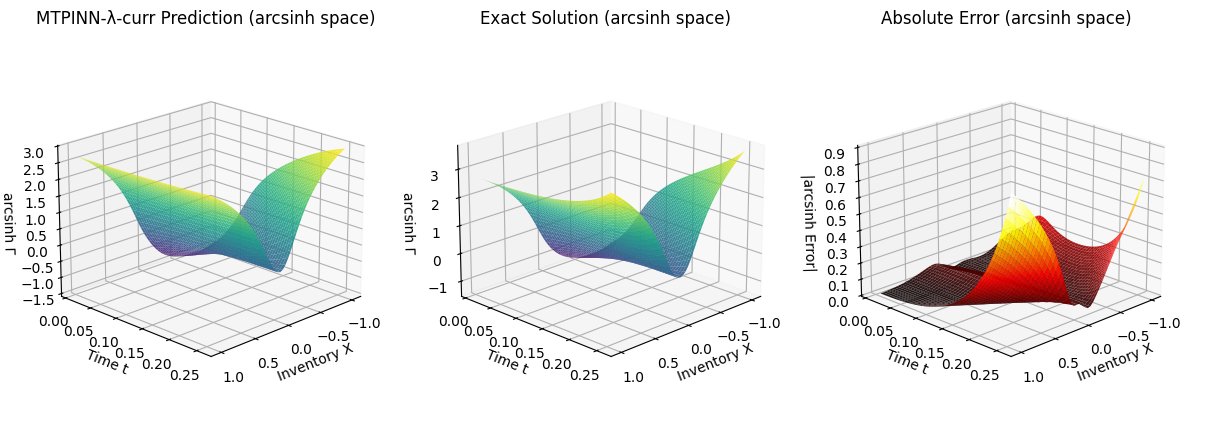}%
    \caption{$\lambda=0.05$}
  \end{subfigure}\hfill
  \begin{subfigure}[t]{0.45\linewidth}
    \centering
    \includegraphics[width=\linewidth]{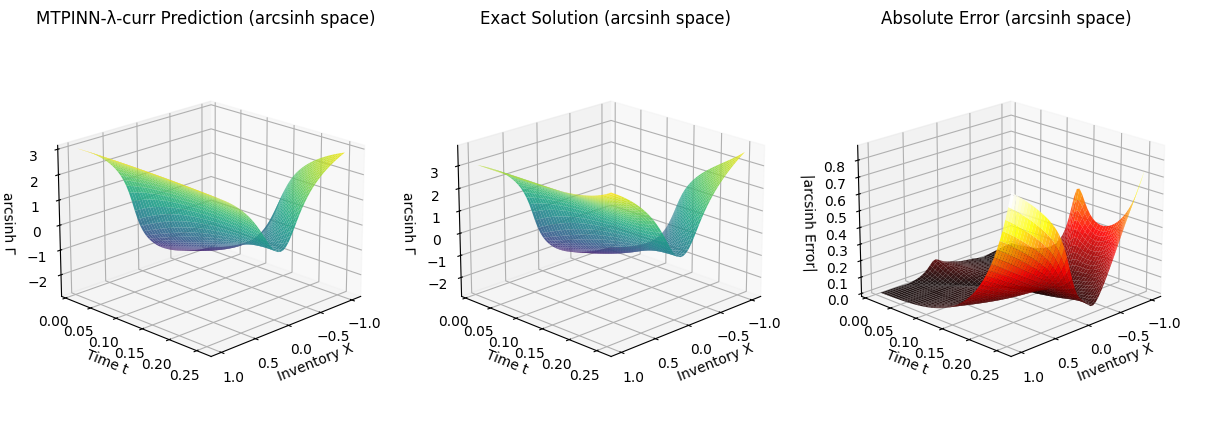}%
    \caption{$\lambda=0.10$}
  \end{subfigure}

  \caption{Arcsinh-space value surfaces for each $\lambda$.}
  \label{fig:value-function-asinh-market}
\end{figure}

\begin{figure}
  \centering
  \begin{subfigure}[t]{1.0\linewidth}
    \centering
    \includegraphics[width=\linewidth]{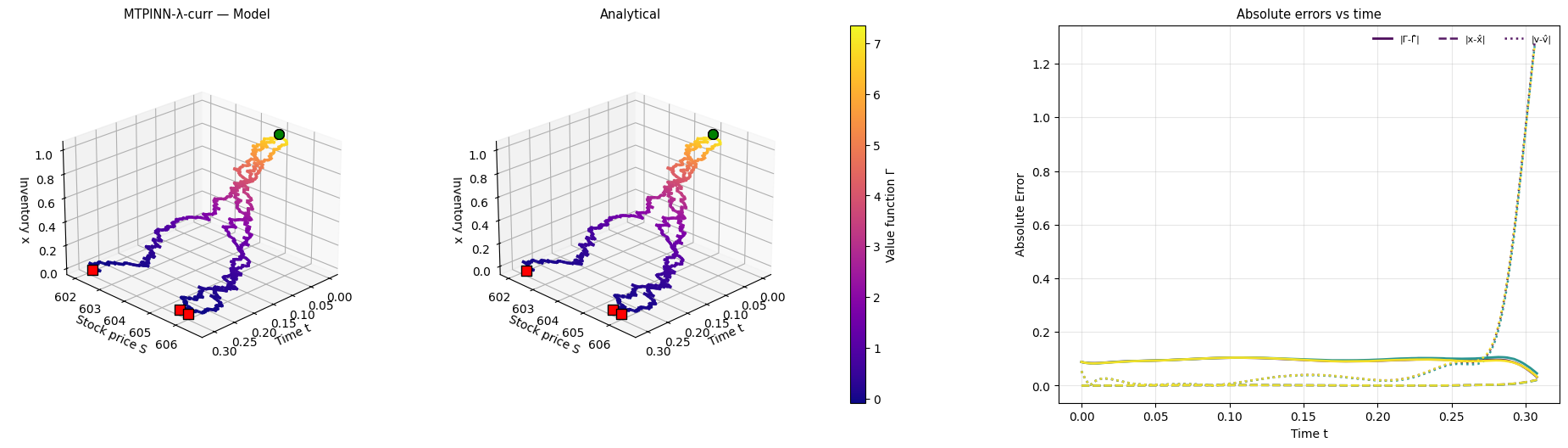}%
    \caption{$\lambda=0.05$}
  \end{subfigure}\hfill
  \begin{subfigure}[t]{1.0\linewidth}
    \centering
    \includegraphics[width=\linewidth]{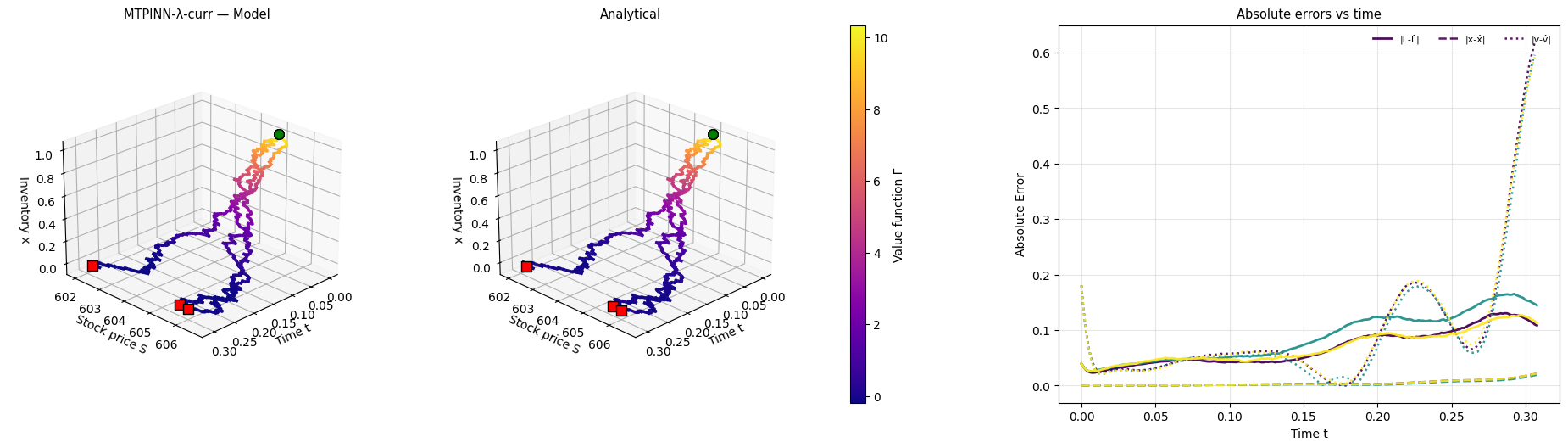}%
    \caption{$\lambda=0.10$}
  \end{subfigure}

  \caption{Pathwise rollouts: model vs. analytic trajectories and absolute error over time for each $\lambda > 0$.}
  \label{fig:paths-market}
\end{figure}

Figure \ref{fig:cost-variance} and Table \ref{tab:cost_variance} both show the cost-variance/exposure trade-offs. It is clear that the MT-PINN is behaving well for varying $\lambda$ in the sense that for higher $\lambda$, the model is experiencing less exposure and for $\lambda=0$ the model behaves very similarly to TWAP, both as expected. For this given SPY market data, TWAP/risk-neutral MT-PINN experience lower costs, likely as a result of rising SPY market prices across the sampled time windows. This is evidenced in Figure \ref{fig:window_overlay}, where the execution trajectories and costs for the MT-PINN with different $\lambda$ and TWAP across four chosen trading windows. The top two time window plots show rising SPY prices and report lowest costs associated with the TWAP algorithm. In contrast, the bottom two time window plots show falling SPY prices and detail lowest costs associated with the highest risk-averse MT-PINN. It is important to mention that this backtest assumed no shorting/over-liquidation and ensures that once zero-inventory is achieved, no subsequent trades are made. 

Figure \ref{fig:final-inventory-dists-market} further exemplifies MT-PINN's consistency with reaching near-zero terminal inventory, especially with lower \(\lambda\) values. Figure \ref{fig:heatmap-market} exhibits the domain-wide inverse hyperbolic sine function absolute value function error (\(\text{arcsinh}(| \Gamma_\text{pred} - \Gamma_\text{exact}|)\)) over \((t, X)\) for fixed stock price \(S=605\), showing low error across most of the domain with the exception to high terminal inventory. This is similarly illustrated in the arcsinh space value surfaces in Figure \ref{fig:value-function-asinh-market}. Nonetheless, Figure \ref{fig:trading-rate-market} and \ref{fig:inventory-traj-market} illustrates the  MT-PINNs ability in closely matching the trading rate and inventory trajectory closed form solution across varying \(\lambda\). Finally, Figure \ref{fig:paths-market} displays the risk-averse MT-PINNs vs analytical inventory trajectories for three paths, showing the value function along the path and the absolute error over time for the value function, inventory trajectory, and trading rate.

\end{document}